\documentclass{article}

\usepackage{PRIMEarxiv}

\usepackage[utf8]{inputenc} 
\usepackage[T1]{fontenc}    
\usepackage{url}            
\usepackage{booktabs}       
\usepackage{amsfonts}       
\usepackage{nicefrac}       
\usepackage{microtype}      
\usepackage{lipsum}
\usepackage{fancyhdr}       
\usepackage{graphicx}       
\graphicspath{{media/}}     

\usepackage[utf8]{inputenc} 
\usepackage[T1]{fontenc}    
\usepackage{url}            
\usepackage{booktabs}       
\usepackage{amsfonts}       
\usepackage{nicefrac}       
\usepackage{microtype}      

\usepackage{subcaption}

\usepackage{microtype}
\usepackage{graphicx}
\usepackage{wrapfig}
\usepackage{caption}
\usepackage{booktabs} 
\usepackage{pifont}

\usepackage{amsmath}
\usepackage[table,xcdraw]{xcolor}
\usepackage{amssymb}
\usepackage{graphicx}
\usepackage{booktabs}
\definecolor{bg}{RGB}{176,226,255}
\definecolor{bonus_green}{RGB}{0,100,0}

\usepackage[colorlinks=true, allcolors=blue]{hyperref}
\definecolor{red}{HTML}{D9423C}
\definecolor{purple}{HTML}{854C98}

\usepackage{amsmath}
\usepackage{amssymb}
\usepackage{mathtools}
\usepackage{amsthm}

\usepackage[capitalize,noabbrev]{cleveref}

\theoremstyle{plain}
\newtheorem{theorem}{Theorem}[section]

\newtheorem{lemma}[theorem]{Lemma}

\theoremstyle{definition}

\theoremstyle{remark}

\usepackage[utf8]{inputenc} 
\usepackage[T1]{fontenc}    
\usepackage{hyperref}       
\usepackage{url}            
\usepackage{booktabs}       
\usepackage{amsfonts}       
\usepackage{nicefrac}       
\usepackage{microtype}      
\usepackage{lipsum}
\usepackage{fancyhdr}       
\usepackage{graphicx}       
\graphicspath{{media/}}     

\usepackage{algorithm}
\usepackage{algorithmic}
\usepackage{makecell}
\usepackage{booktabs} 
\usepackage{threeparttable}

\usepackage{multirow}
\usepackage[most]{tcolorbox}
\usepackage{float}
\definecolor{c1}{HTML}{F898CB}
\definecolor{c2}{HTML}{D9DEE7}
\definecolor{color_pl}{HTML}{7AC3DF}
\definecolor{sys_prompt}{HTML}{018B38}
\definecolor{question}{HTML}{652884}
\definecolor{CoT}{HTML}{F5A216}
\definecolor{answer}{HTML}{6A8EC9}

\title{DGRO: Enhancing LLM Reasoning via Exploration-Exploitation Control and Reward Variance Management
}

\author{
  Xuerui Su \thanks{~~These authors contributed equally to this work.} \\
School of Mathematics and Statistics \\
Beijing Jiaotong University \\
\texttt{24110486@bjtu.edu.cn} \\
\And
Liya Guo \footnotemark[1]\\
Yau Mathematical Sciences Center, Tsinghua University\\
Department of Mathematical Sciences,  Tsinghua University\\
Beijing, China\\
 \texttt{gly22@mails.tsinghua.edu.cn}\\
\And
Yue Wang \\
Zhongguancun Academy \\
\texttt{yuewang\_yw@foxmail.com} \\
\And
Yi Zhu\\
Yau Mathematical Sciences Center, Tsinghua University\\
Yanqi Lake Beijing Institute of Mathematical Sciences and Applications\\ Beijing, China\\
\texttt{yizhu@tsinghua.edu.cn}
   \And
  Zhiming Ma \\
  Academy of Mathematics and Systems Science \\
  Chinese Academy of Sciences\\
  \texttt{mazm@amt.ac.cn} \\
  \And
Zun Wang \footnotemark[2]\\
Microsoft Research AI4Science\\
 Beijing, China.\\	
 \texttt{zunwang@microsoft.com} \\
\And
Yuting Liu\thanks{~~Corresponding author.} \\
School of Mathematics and Statistics \\
Beijing Jiaotong University \\
\texttt{ytliu@bjtu.edu.cn} \\
}

\begin{document}
\maketitle

\begin{abstract}
Inference scaling further accelerates Large Language Models (LLMs) toward Artificial General Intelligence (AGI), with large-scale Reinforcement Learning (RL) to unleash long Chain-of-Thought reasoning. Most contemporary reasoning approaches usually rely on handcrafted rule-based reward functions. However, the tarde-offs of exploration and exploitation in RL algorithms involves multiple complex considerations, and the theoretical and empirical impacts of manually designed reward functions remain insufficiently explored. In this paper, we propose Decoupled Group Reward Optimization (DGRO), a general RL algorithm for LLM reasoning. On the one hand, DGRO decouples the traditional regularization coefficient into two independent hyperparameters: one scales the policy gradient term, and the other regulates the distance from the sampling policy. This decoupling not only enables precise control over balancing exploration and exploitation, but also can be seamlessly extended to Online Policy Mirror Descent (OPMD) algorithms in Kimi k1.5 and Direct Reward Optimization. On the other hand, we observe that reward variance significantly affects both convergence speed and final model performance. We conduct both theoretical analysis and extensive empirical validation to assess DGRO, including a detailed ablation study that investigates its performance and optimization dynamics. Experimental results show that DGRO achieves state-of-the-art performance on the Logic dataset with an average accuracy of 96.9\%, and demonstrates strong generalization across mathematical benchmarks.
\end{abstract}


\section{Introduction}
\label{sec:intro}
Inference scaling has significantly enhanced the reasoning capabilities of Large Language Models (LLMs) \cite{ouyang2022training,grattafiori2024llama,claude3,DeepSeekV3}, as exemplified by OpenAI O1 \cite{jaech2024openai} and DeepSeek R1 \cite{DeepSeekR1}. A key factor in this progress is large-scale reinforcement learning (RL) \cite{sutton1998reinforcement}, which plays a central role in optimizing LLMs for complex reasoning tasks \cite{hou2025advancing,jin2025search,zhang2025r1}. Modern RL algorithms like Proximal Policy Optimization (PPO) \cite{PPO} often seek to maximize task-specific rewards while constraining the divergence from a reference policy, balancing performance and stability. Building upon PPO, Group Relative Policy Optimization (GRPO) \cite{shao2024deepseekmath} employs group-wise normalized rewards as advantages, and demonstrates strong performance on reasoning benchmarks \cite{DeepSeekR1}. 
Alternatively, reward optimization algorithms that directly minimize the deviation between rewards and policy log-probabilities, such as Kimi k1.5 \cite{team2025kimi} for LLM reasoning and Direct Reward Optimization (DRO) \cite{DRO_Deepmind} for aligning agent behavior with human preferences, offer another promising direction for RL research.

Despite this progress, the study of how the reward function interacts with the optimization mechanism in reasoning tasks remains limited. In typical Reinforcement Learning from Human Feedback (RLHF), a reward model is trained to serve as a reward signal at later stages of RL training \cite{ouyang2022training,deep_rlhf}. Alternatively, recent works such as DeepSeek R1 \cite{DeepSeekR1}, Kimi k1.5 \cite{team2025kimi}, and Logic-RL \cite{xie2025logic} adopt rule-based reward functions to reduce training costs and improve generality. Although these approaches are practical, reward designs are often heuristic and lack theoretical grounding on the properties of reward functions that lead to more effective RL optimization. Moreover, the impact of RL algorithms on the exploration–exploitation trade-off in reasoning tasks remains underexplored and theoretically ungrounded.

In this paper, we introduce Decoupled Group Reward Optimization (DGRO), a general RL algorithm for LLM reasoning. Specifically, DGRO decouples the regularization coefficient in the KL-regularized reward maximization objective into two independent coefficients: one controlling the strength of the policy gradient, and the other governing the distance between reference policy and actor policy. This decoupling provides finer control over both optimization dynamics and reasoning quality, enabling a more effective balance between exploration and exploitation. 
Theoretically, we provide theoretical insights about how the decoupled coefficients and reward variance influence reasoning performance and optimization dynamics. Moreover, we also analyze the validity and error bound of using the reward mean as an approximation for the soft value function under various conditions. These findings provide practical guidance for exploration-exploitation control and reward variance management. Our theoretical predictions are validated through experiments on the K\&K and Math datasets, where DGRO achieves superior performance compared to the base model. Our main contributions are summarized as follows:
\begin{enumerate}
    \item We propose DGRO, a general reward optimization algorithm for LLM reasoning tasks. The proposed objective extends prior methods and enables more efficient exploration–exploitation trade-offs through the decoupling of regularization coefficients.
    \item We provide the theoretical analysis about the effect of the decoupled coefficients and reward variance. Appropriate combinations of decoupled coefficients and higher reward variance can facilitate faster convergence and better final accuracy.
    \item We demonstrate the effectiveness of DGRO on reasoning benchmarks like Logic and Math tasks. Ablation studies further support our theoretical claims and offer practical guidance on exploration-exploitation control and reward variance management.
\end{enumerate}

\section{Background}
\label{sec:preliminary}

\subsection{Related Works}
RL algorithms have been widely used to improve the reasoning and human alignment abilities of LLMs \cite{DeepSeekR1,xie2025logic,DAPO,lin2025cppo,Dr.GRPO,DPO,su2025reveal}. Among various approaches, \emph{Proximal Policy Optimization (PPO)}~\cite{schulman2017proximal, PPO-basedRLHF} remains a widely used RL algorithm, typically via a KL-regularized reward maximization objective. Building upon this, DeepSeek R1 \cite{DeepSeekR1} adopts \emph{Group Relative Policy Optimization (GRPO)}~\cite{shao2024deepseekmath}, which employs group-wise normalized rewards as advantages, thereby improving sample efficiency and reducing gradient variance.
Rather than relying on PPO’s surrogate objective \cite{PPO-basedRLHF}, an alternative line of research reformulates the reward optimization problem itself, deriving exact closed-form updates and designing new reinforcement learning algorithms accordingly that directly minimize the deviation between rewards and policy log-probabilities \cite{DRO_Deepmind,su2025reveal,team2025kimi}. For instance, \emph{Direct Reward Optimization (DRO)} \cite{DRO_Deepmind} directly approximates the reward-maximizing policy in Human Alignment tasks. And similar approach has also been used for enhancing LLM reasoning ability like \emph{Kimi k1.5} \cite{team2025kimi}, but without providing systematic theoretical analysis or guarantees.

Reward modeling is important for reasoning ability. Traditional RL pipelines train reward models from data, but these are susceptible to reward hacking~\cite{everitt2017reinforcement, gao2023scaling}. As a result, preference-based approaches such as \emph{Direct Preference Optimization (DPO)}~\cite{DPO}, \emph{Identity Preference Optimization (IPO)}~\cite{IPO}, and rule-based reward assignment~\cite{DeepSeekR1, DAPO, xie2025logic} are introduced. Rule-based methods typically assign predefined rewards (e.g., $+1$ for correct reasoning chains and $-1$ otherwise).
In addition to reward modeling, other training parameters such as response length and regularization strength have also been found to affect the reasoning quality of RL-trained models~\cite{tang2024generalized, aggarwal2025l1}. 
While these methods have led to significant empirical progress, most lack theoretical analysis regarding how reward formulation and optimization dynamics influence model behavior. This motivates a more principled study of reward-based RL for reasoning tasks.

\subsection{Preliminary}\label{sec:preliminary_2}
In the response reasoning process, LLM is prompted with an input question $x$ to generate an output $y$, where both $x,y$ are sequences of tokens. The goal of a reasoning model is to learn a chain-of-thought policy that produces accurate responses through structured reasoning. PPO \cite{PPO-basedRLHF} formulates the objective as maximizing the expectation of reward $r_{\phi}$ while penalizing deviation from a source policy $\pi_{\text{src}}$. This is typically done using a KL-regularized objective \cite{schulman2017equivalence,schulman2017proximal,DRO_Deepmind}:
\begin{equation}
\label{eq:r_kl}
\mathbb{E}_{x \sim D, y \sim \pi_\theta(\cdot \mid x)}\left[r_{\phi}(x, y)\right] - \beta D_{\mathrm{KL}}\left(\pi_\theta(\cdot \mid x) \| \pi_{\text{src}}(\cdot \mid x)\right),
\end{equation}
where the regularization coefficient $\beta$ controls the regularization strength, and $r_{\phi}$ is a trained reward model. $\pi_{\text{src}}(\cdot|\cdot)$ is the source policy, which may refer to the reference policy $\pi_{\text{ref}}(\cdot|\cdot)$ corresponding to the PPO algorithm \cite{PPO-basedRLHF} or old policy $\pi_{\theta_{\text{old}}}(\cdot|\cdot)$ optimized by the last round corresponding to the Policy Mirror Descent (PMD) algorithm \cite{tomar2020mirror,team2025kimi}, depending on different implementation. The optimal solution to the KL-constrained reward maximization objective in Eq.\eqref{eq:r_kl} is:
\begin{equation}
    \begin{aligned}
        \label{eq:r_analytic}
        r^*(x, y) = V(x, \beta) + \beta \log\left(\frac{\pi_\theta(y \mid x)}{\pi_{\text{src}}(y \mid x)}\right),
    \end{aligned}
\end{equation}
where the soft value function $V(x, \beta) = \beta \log Z(x)$, following the convention (Eq.5) of \cite{haarnoja2017reinforcement},
and the partition function $Z(x)=\sum_y \pi_{\text{src}}(y \mid x) \exp \left(\frac{1}{\beta} r(x, y)\right)$. Different from PPO \cite{PPO-basedRLHF}, another methodology \cite{richemond2024offline,team2025kimi} 
directly minimizes the deviation between rewards and policy log-probabilities based on Eq.\eqref{eq:r_analytic}. We extends this methodology to the following loss function:
\begin{equation}
\label{eq:GRO_loss}
{  \begin{aligned}
\mathcal{L}(\theta) &= \mathbb{E}_{x \sim \mathcal{D},y \sim \pi_{\text{sam}}(\cdot|x)} \left[\left(r(x, y) - V(x, \beta)
- \beta \log \frac{\pi_\theta\left(y \mid x\right)}{\pi_{\text{src}}\left(y \mid x\right)}\right)^2
\right]\,,
\end{aligned}}
\end{equation}
where $\pi_{\text{sam}}(\cdot|\cdot)$ is the sampling policy and $\pi_{\text{src}}(\cdot|\cdot)$ is the source policy, which may refer to the old policy $\pi_{\theta_{\text{old}}}(\cdot|\cdot)$ or reference policy $\pi_{\text{ref}}(\cdot|\cdot)$ depending on different implementation.

\section{Decoupled Group Reward Optimization algorithm}\label{sec:method}


To decouple the multiple effects of the regularization coefficient $\beta$ in Eq.\eqref{eq:GRO_loss}, we propose Decoupled Group Reward Optimization (DGRO), a generalized reward-based optimization algorithm built upon Eq.\eqref{eq:GRO_loss} mentioned in Section \ref{sec:preliminary_2}.
\begin{equation}
\label{eq:dgro_loss}
{  \begin{aligned}
\mathcal{L}_{\text{DGRO}}(\theta) &= -\mathbb{E}_{x \sim \mathcal{D},y \sim \pi_{{\text{sam}}}(\cdot|x)} \left[\beta_1
\left( r(x, y) - V(x,\beta) \right)\log \pi_\theta(y \mid x) 
- \beta_2\left(\log \frac{\pi_\theta\left(y \mid x\right)}{\pi_{\text{src}}\left(y \mid x\right)}\right)^2
\right]\,,
\end{aligned}}
\end{equation}
where $\beta_1$ determines the influence of the policy gradient term, and $\beta_2$ governs the strength of the policy distance penalty. The gradient of the loss~\eqref{eq:dgro_loss} can be estimated by:
\begin{equation}\label{eq:grad_loss}
    \begin{aligned}
-\beta_1 \left( r(x, y) -V(x,\beta) \right)\nabla_\theta \log \pi_\theta(y \mid x) 
+ \beta_2 \nabla_\theta\left(\log \frac{\pi_\theta\left(y \mid x\right)}{\pi_{{\text{src}}}\left(y \mid x\right)}\right)^2. 
\end{aligned}
\end{equation}

This objective generalizes the method proposed in Ref. \cite{team2025kimi,DRO_Deepmind}, which can be seen as a special case of DGRO from the perspective of algorithm, as shown in Table \ref{tab:DGRO_and_others}. Due to space limitation, we detail the comparison with other works in Appendix \ref{appendix:DGRO_and_others}. 

\begin{table}[ht]
\centering
\vspace{-0.1in}
\caption{Summary of DGRO and other related work.}
\renewcommand{\arraystretch}{1.2}
\begin{threeparttable} 
\resizebox{0.9\hsize}{!}{
\begin{tabular}{lcccccc}
\toprule
\textbf{Algorithm} & \makecell{\textbf{Policy-Gradient} \\ \textbf{Coefficient} $\beta_1$} & \makecell{\textbf{Policy-Distance} \\ \textbf{Coefficient} $\beta_2$} & \makecell{$V(x,\beta)$} & \makecell{\textbf{Sampling Policy} \\$\pi_{sam}$} & \makecell{\textbf{Source Policy} \\$\pi_{src}$} & \makecell{\textbf{Offline /} \\ \textbf{Online}} \\
\midrule
DRO \cite{DRO_Deepmind} & $\beta$ & $\frac{\beta^2}{2}$ & $V_\phi(x)$\tnote{1} & $\mu(\cdot|x)$\tnote{2} & $\pi_{\text{ref}}$ & Offline  \\
Kimi k1.5 \cite{team2025kimi} \tnote{3} & 1 & $\frac{\beta}{2}$ &  $\bar{r}(x)$ & $\pi_{\theta_{\text{old}}}$ & $\pi_{\theta_{\text{old}}}$ & Online  \\
\textbf{DGRO} & $\beta_1$ & $\beta_2$ &  $\bar{r}(x)$ & $\pi_{\theta_{\text{old}}}$ & $\pi_{\theta_{\text{old}}}$ & Online  \\
\bottomrule
\end{tabular}
}
\begin{tablenotes}
  \footnotesize  
     \item[1] $V_\phi(x)$ is parameterized by a NN with parameters $\phi$.
     \item[2] $\mu(\cdot|x)$ is a fixed sampling policy.
     \item[3] serves as a variant of Online Policy Mirror Descent (OPMD) algorithm.
   \end{tablenotes}
   \end{threeparttable}
\label{tab:DGRO_and_others}
\vspace{-0.1in}
\end{table}

Comparing with purely offline approaches, online reinforcement learning where updates are performed with data generated by the recent policy, naturally reduces distribution shift \cite{panaganti2023bridging}, improves sample efficiency, and stabilizes training. To harness these benefits in DGRO, we therefore adopt an online setting by letting both the sampling policy and the source policy be the previous iteration’s policy, i.e. $\pi_{\text{sam}} = \pi_{\text{src}} = \pi_{\theta_{\text{old}}}$ (see Table \ref{tab:DGRO_and_others}). This choice ensures that each gradient estimate of the DGRO loss remains grounded in the most recent behavior policy, mitigating distribution shift and yielding more reliable updates. While this setting is technically off-policy, it is commonly referred to as "online" in practice due to the short lag between data collection and policy updates.


To understand the optimization behavior of DGRO, we provide a two-part theoretical analysis. First, since computing $V(x,\beta)$ is intractable due to the partition function $Z(x)$, we introduce an approximation strategy and derive an upper bound on the resulting error (Section \ref{sec:error_approx}). Second, we analyze how the regularization coefficients $(\beta_1, \beta_2)$ influence convergence behavior, showing that reward variance is a key factor affecting convergence speed (Section \ref{sec:gradient_loss}).

\subsection{Theoretical analysis for using $\bar{r}(x)$ to approximate $V(x,\beta)$}
\label{sec:error_approx}

Computing $V(x,\beta)$ is often intractable due to the partition function $Z(x)$. Ref. \cite{DRO_Deepmind} addresses this by training an auxiliary model to approximate the full expression, which introduces both modeling complexity and potential bias. Alternatively, Ref. \cite{team2025kimi} uses the group-wise average reward $\bar{r}(x)$ as a surrogate for $V(x,\beta)$, though without formal justification. In this section, we provide a theoretical analysis supporting the approximation $\bar{r}(x) \approx V(x,\beta)$. Based on group-wise rewards, the DGRO gradient then becomes:
\begin{equation}
\label{eq:grad_loss_mean}
    \begin{aligned}
-\frac{1}{G} \sum_{j=1}^{G} \left(\beta_1
\left( r(x, y_i) - \bar{r}(x) \right)\nabla_\theta \log \pi_\theta(y_i \mid x) 
- \beta_2 \nabla_\theta\left(\log \frac{\pi_\theta\left(y_i \mid x\right)}{\pi_{\theta_{\text{old}}}\left(y_i \mid x\right)}\right)^2
\right)\,,
\end{aligned}
\end{equation}
where $G$ denotes the rollout times, and $\bar{r}(x)$ is the mean reward across the sampled responses.
While $\beta$ is decoupled into $\beta_1$ and $\beta_2$, to simplify the proof, we still use $\beta$ for $V(x, \beta)$ in the following theoretical analysis. Specifically, we derive explicit error bounds in the two asymptotic regimes $\beta\to 0$ and $\beta\to\infty$, thereby clarifying the conditions under which this approximation is valid.
\begin{lemma}
\label{lemma:unbiased_estimation}
Denote $\bar{r}(x)=\mathbb{E}_{y\sim \pi_{{\text{sam}}}(\cdot|x)}[r(x,y)]$, $V(x, \beta)=\beta\log\sum_y \pi_{\text{src}}(y \mid x) \exp \left(\frac{1}{\beta} r(x, y)\right)$. When $\pi_{{\text{sam}}}=\pi_{{\text{src}}}$, the following equation holds if and only if $\beta\to\infty$ or $r(x,y)=c(x)$ holds almost everywhere under measure $\pi_{{\text{sam}}}(\cdot|x)$, where $c(x)$ depends only on $x$:
\begin{equation}
    \bar{r}(x) = V(x,\beta).
\end{equation}
\end{lemma}
See proof in Appendix \ref{proof:unbiased_estimation}. For DGRO implementation, $\pi_{\text{sam}} = \pi_{\text{src}} = \pi_{\theta_{\text{old}}}$ as mentioned before. In the following content, we adopt the setting of $\pi_{\text{sam}} = \pi_{\text{src}}$ by default.  Lemma \ref{lemma:unbiased_estimation} provides theoretical support for approximating $V(x, \beta)$ by $\bar{r}(x)$ as $\beta \to \infty$, as also used in Ref. \cite{team2025kimi}. However, in practice, $\beta$ is typically set to a small value, and recent work \cite{DAPO} even reports improved empirical performance when the KL regularization is omitted (i.e., $\beta = 0$). This raises concerns about the validity of the approximation in low-$\beta$ regimes. To address this, we propose Lemma \ref{lemma:biased_estimation}, which characterizes the approximation error between $\bar{r}(x)$ and $\beta \log Z(x)$ as $\beta \to 0$. 

\begin{lemma}
\label{lemma:biased_estimation}
Denote $m(x)=\sup _y r(x, y)$. Assume $\pi_{{\text{sam}}}=\pi_{{\text{src}}}$. In the limit $\beta \to 0$, a non-zero error exists between $\bar{r}(x)$ and $\beta \log\left(Z(x)\right)$, if the reward function is not constant:
\begin{equation}
\lim_{\beta\to0}V(x,\beta) - \bar{r}(x) = m(x)-\bar{r}(x).
\end{equation}
\end{lemma}
If the reward function is non-constant, then the error $m(x) - \bar{r}(x)$ is strictly positive, indicating that the approximation $\bar{r}(x) \approx \beta \log Z(x)$ incurs a non-negligible error when $\beta$ is small (see the proof of Lemma \ref{lemma:biased_estimation} in Appendix \ref{proof:biased_estimation}). 
To further investigate how this error depends on $\beta$ as it approaches zero, we provide a more refined error upper bound in Theorem \ref{theorem:error_of_taulogz-small_tau}.
\begin{theorem}
\label{theorem:error_of_taulogz-small_tau}
Denote $p_*(x)=P\left(r(x, y)=\sup _y r(x, y)\right)$, $m(x)=\sup _y r(x, y)$. Assume that $\pi_{{\text{sam}}}=\pi_{{\text{src}}}$. We have:
    \begin{equation}
0\leq V(x,\beta) - \bar{r}(x) \leq m(x)-\bar r(x) + \beta\left(\log p_*(x) + \frac{1-p_*(x)}{p_*(x)}\right).
    \end{equation} 
\end{theorem}
See the proof of Theorem \ref{theorem:error_of_taulogz-small_tau} in Appendix \ref{proof:error_of_taulogz-small_tau}. The above result shows that the approximation error is bounded and diminishes as $\beta$ decreases (as $\log p_*(x) + \frac{1-p_*(x)}{p_*(x)}\geq 0$, see proof in Appendix \ref{proof:beta_coef_big_0}). While the approximation incurs a nonzero error as $\beta \to 0$, it remains effective in practice. This is likely because the approximate objective yields a similar optimization direction, especially as decoupling $\beta$ into $\beta_1$ and $\beta_2$ helps control the error (with straightforward derivation, $\beta = {2\beta_2}/{\beta_1}$; see Eq.\ref{eq:DGRO_beta_corresponding} in Appendix \ref{appendix:DGRO_and_others}). Additionally, managing reward variance and the fact that $m(x) - \bar{r}(x)$ vanishes as the policy improves further mitigate the impact. Therefore, despite the theoretical gap in the small-$\beta$ regime, DGRO still achieves strong empirical performance. 



\subsection{Gradient Behavior under Varying Reward Variance and Coefficient Settings}
\label{sec:gradient_loss}
Both reward variance and the decoupled coefficients $\beta_1$ and $\beta_2$ play a critical role in shaping the learning dynamics. We further analyze their impact on optimizing the DGRO objective. Theorem \ref{theorem:var} characterizes how the magnitude of the policy gradient is influenced by reward variance and how $\beta_1$ and $\beta_2$ individually affect the gradient of DGRO.
\begin{theorem}
    \label{theorem:var}
    Denote $\bar{r}(x)=\mathbb{E}_{y\sim \pi_{\theta_{\text{old}}}(\cdot|x)}[r(x,y)]$. For $\nabla_\theta \mathcal{L}_{\text{DGRO}}(\theta)$, we have:
    \begin{equation}
    \begin{aligned}
        \nabla_\theta \mathcal{L}_{\text{DGRO}}(\theta) = &-\mathbb{E}_{x \sim \mathcal{D}}\underbrace{\mathbb{E}_{y\sim \pi_{\theta_{\text{old}}}(\cdot|x)}\left[\beta_1\left(r\left(x, y\right)-\bar{r}(x)\right)\nabla_\theta \log \pi_\theta\left(y \mid x\right)\right]}_{\text{PG}(x,\theta)}\\
        &\quad\quad\quad\quad\quad-\mathbb{E}_{x \sim \mathcal{D}}\underbrace{\mathbb{E}_{y\sim \pi_{\theta_{\text{old}}}(\cdot|x)}\left[{\beta_2} \nabla_\theta\left(\log \frac{\pi_\theta\left(y \mid x\right)}{\pi_{\theta_{\text{old}}}\left(y \mid x\right)}\right)^2\right]}_{\text{Normaliser}(x,\theta)}.
    \end{aligned}
    \end{equation}
    Denote $\operatorname{Var}_{y\sim \pi_{\theta_{\text{old}}}(\cdot|x)}[r(x,y)]=\mathbb{E}_{y\sim \pi_{\theta_{\text{old}}}(\cdot|x)}[(r(x,y)-\bar{r}(x))^2]$. Assume that $r(x,y)$ is bounded by $r_{\text{max}}$ and has finite variance, $|\log\frac{\pi_\theta(y|x)}{\pi_{\theta_{\text{old}}}(y|x)}| \leq \eta$ and $\pi_\theta(y|x)$ is parameterized by $\text{Softmax}(f({x},{y};\theta))$, the following property holds:
    \begin{equation}
    \begin{aligned}
        &\left\|\text{PG}(x,\theta)\right\| \leq  (4r_{\text{max}}+2)N\beta_1\gamma(x ; \theta)\operatorname{Var}_{y \sim \pi_{\theta_{\text{old}}}(\cdot \mid x)}[r(x, y)]^{\frac{1}{3}}\\
        &\left\|\text{Normaliser}(x,\theta)\right\|\leq 4N\eta\beta_2\gamma(x ; \theta),
    \end{aligned}
    \end{equation}
    where $\gamma(x;\theta):=\max_{y \in \mathcal{Y},\ n \in {1, \ldots, N}} || J_{f(x, y_{\leq n-1}; \theta)} ||_2$, with $y_{\leq n-1}$ denoting the first $n$ tokens of the response $y$ and $J_{f({x}, {y}_{\leq n-1} ; \theta)}$ is the Jacobian of $f\left({x}, {y}_{\leq n-1}; \theta\right)$ with respect to $\theta$, and $||\cdot||$ and $||\cdot||_2$ denote the Euclidean and operator norms, respectively.
\end{theorem}
See the proof of Theorem \ref{theorem:var} in Appendix \ref{proof:var}. Theorem~\ref{theorem:var} suggests that lower reward variance leads to smaller gradients, which may slow down the optimization process. Thus, in practice, reward designs should avoid overly small variance to maintain sufficient learning signals.
Moreover, Theorem~\ref{theorem:var} implies that $\beta_1$ controls the strength of exploration by scaling the policy gradient term, while $\beta_2$ regulates exploitation by penalizing deviations from the old policy. Therefore, $\beta_1$ and $\beta_2$ should be adaptively tuned in practice to balance the trade-off between exploration and exploitation.

\section{Experiment}
\label{sec:exp}
\subsection{DGRO Implementations}
In our subsequent experiments, DGRO considers a rule-based reward model that we do not train a separate reward model. Instead, rewards are directly assigned based on predefined rules tailored to different response types. In this section, we detail the implementation of DGRO, including the design of reward rules, the computation of the group-based loss function, and the policy update procedure.

To encourage the model to develop human-like chains of reasoning, DGRO enforces structural constraints on the model’s output format. Specifically, the reasoning steps should be enclosed within <think> and </think> tags, and final answers within <answer> and </answer> tags, following practices from previous works \cite{DeepSeekR1, su2025trust}. Any response that violates this format is considered invalid during training.
We assume the total reward $r$ is a weighted sum of three components:
\begin{equation}
    \begin{aligned}
    \label{eq:reward}
        r = c_f \cdot r_f + c_{\text{corr}} \cdot r_{\text{corr}} + c_{\text{com}} \cdot r_{\text{com}}\,,
    \end{aligned}
\end{equation}
where \( r_f \), \( r_{\text{corr}} \), and \( r_{\text{com}} \) represent binary scores (in $\{-1,+1\}$) for format correctness, answer correctness, and answer completeness, respectively. The coefficients \( c_f \), \( c_{\text{corr}} \), and \( c_{\text{com}} \) are the corresponding weights. For instance, one practical configuration is \( c_f = 1 \), \( c_{\text{corr}} = 1.75 \), and \( c_{\text{com}} = 0.25 \). Given this reward structure, responses receive different scores depending on their quality:
\begin{itemize}
    \item Correct format, correct answer, and complete response: \( r = 1 + 1.75 + 0.25 = 3 \)
    \item Correct format but incorrect answer: \( r = 1 - 1.75 + 0.25 = -0.5 \)
    \item Correct format but incomplete answer: \( r = 1 - 1.75 - 0.25 = -1 \)
    \item Incorrect format (e.g., missing required tags): \( r = -1 - 1.75 - 0.25 = -3 \)
\end{itemize}
In the following experiments, we control reward variance by adjusting the reward range. Based on the predefined reward rules, DGRO employs a group-wise optimization strategy: for each input prompt $x$, the model generates multiple candidate responses (i.e., rollouts) $\{y_j\}_{j=1}^G$, which are treated as a group. Each candidate in the group is individually scored using the rule-based reward model. The average reward across the group serves as a reference signal $\bar{r} = \frac{1}{G}\sum_{j=1}^G r(y_j, x)$ to calculate the loss. DGRO then optimizes the model to increase the relative reward of high-quality responses within the group, promoting both learning stability and reward sensitivity.
The training procedure is summarized in Algorithm~\ref{alg}, and concrete examples are provided in Appendix~\ref{sec:case_study}.

\begin{algorithm}
\caption{DGRO}
\begin{algorithmic}[1]
\label{alg}
\REQUIRE Prompt dataset $\mathcal{D}_{\text{raw}}$, base model $\pi_{\text{ref}}$.
\REQUIRE Reward rules and parameters of designed format, correct, and complete score in Eq.\eqref{eq:reward}.
\REQUIRE Policy gradient coefficient $\beta_1$, policy distance coefficient $\beta_2$, rollout number $G$.
\STATE Initialize actor policy model $\pi_{\theta}$ from the base model $\pi_{\text{ref}}$; $\pi_{\theta_\text{old}} = \pi_{\theta}$.
\FOR{each epoch}
    \STATE Sample a batch of prompts $\{x\}$ from $\mathcal{D}_{\text{raw}}$.
    \STATE For each prompt $x$, generate a group of $G$ responses $\{y_i\}_{i=1}^G$ from $\pi_{\theta_{\text{old}}}(y \mid x)$.
    \STATE Compute rewards $\{r_i\}_{i=1}^G$ by Eq.\eqref{eq:reward}, and calculate the mean reward $\bar{r}(x)$ for each group.
    \STATE Compute policy gradient loss Eq.\eqref{eq:grad_loss_mean}. Update $\pi_\theta$ by descending Eq.\eqref{eq:grad_loss_mean}. 
    \STATE $\pi_{\theta_\text{old}} \leftarrow \pi_{\theta}$.
\ENDFOR
\STATE Save the final policy model.
\end{algorithmic}
\end{algorithm}

\subsection{Experimental Results}\label{sec:Experimental_Results}

In this section, we present the experimental results of DGRO on the K\&K logic puzzle dataset \cite{xie2024memorization} and the Math dataset \cite{deepscaler2025,hu2025orz}. 
For the K\&K logic puzzle dataset, the model is trained for $7{,}200$ steps on 4$\times$A100-80GB GPUs, using a constant learning rate of $4\times10^{-7}$, a maximum response length of $4{,}096$ tokens, and a temperature of $1.0$. To ensure fair comparison, we match the total training data volume---defined as \texttt{batch\_size} $\times$ \texttt{rollout\_times} $\times$ \texttt{training\_steps}---with that of Logic-RL~\cite{xie2025logic}. 
For the math reasoning tasks, we use a fixed learning rate of $4\times10^{-7}$, a batch size of $8$, and rollout times $8$. Additional hyperparameters are listed in Table~\ref{tab:hyperpameters} in Appendix \ref{sec:exp_details}.

We evaluate model performance using the pass@$1$ metric. To gain deeper insight into training dynamics and model behavior, we report three additional metrics in ablation study: (1) \texttt{Validation Accuracy} is measured every 100 steps to track the overall correctness in training. (2) \texttt{Average Response Length} is tracked to measure the length of generated reasoning chains and answers. (3) \texttt{Entropy Loss} is tracked to assess response stability and exploration behavior during training.

\begin{table}[h]
\centering
\caption{\centering Evaluation of reasoning and base models on the K\&K logic puzzle dataset. All scores are pass$@1$ results.}
\label{tab:eval_logic}
\resizebox{0.95\linewidth}{!}{
\begin{tabular}{lccccccccl}
\toprule
\textbf{Model} & 3 & 4 & 5 & 6 & 7 & 2 (OOD) & 8 (OOD) & \textbf{Avg.} \\
\midrule
o3-mini-high & 0.98 & 0.97 & \underline{0.95} & 0.94 & 0.89 & \underline{0.99} & 0.83 & 0.935 \\
o1-2024-12-17 & 0.51 & 0.38 & 0.38 & 0.35 & 0.30 & 0.83 & 0.20 & 0.421 \\
Deepseek-R1 & 0.73 & 0.77 & 0.78 & 0.75 & 0.88 & 0.91 & 0.83 & 0.807 \\
\midrule
GPT-4o & 0.57 & 0.49 & 0.32 & 0.23 & 0.21 & 0.68 & 0.11 & 0.373 \\
GPT-4o-mini & 0.42 & 0.34 & 0.17 & 0.09 & 0.10 & 0.63 & 0.01 & 0.251 \\
NuminaMath-7B-CoT & 0.13 & 0.12 & 0.05 & 0.01 & 0.00 & 0.28 & 0.00 & 0.084 \\
Deepseek-Math-7B & 0.21 & 0.08 & 0.06 & 0.02 & 0.00 & 0.35 & 0.00 & 0.103 \\
Qwen2.5-Base-7B & 0.34 & 0.16 & 0.09 & 0.01 & 0.00 & 0.41 & 0.00 & 0.143 \\
\midrule
Qwen2.5-7B-Instruct-1M & 0.40 & 0.25 & 0.11 & 0.06 & 0.02 & 0.49 & 0.01 & 0.191 \\
+Logic-RL & \underline{0.99} & 0.94 & 0.92 & \underline{0.91} & 0.80 & \underline{0.99} & 0.67 & 0.889 \\
\rowcolor{bg!70}\textbf{+DGRO}, $r\in [-3,3]$ & 0.98 & \underline{0.99} & \underline{0.95} & 0.89 & \underline{0.90} & 0.96 & \underline{0.82} & \underline{0.927} \\
\rowcolor{bg!70}\textbf{+DGRO}, $r\in [-5,5]$ & \textbf{1.0} & \textbf{1.0} & \textbf{0.98} & \textbf{0.98} & \textbf{0.95} & \textbf{1.0} & \textbf{0.87} & \textbf{0.969} \\
\bottomrule
\end{tabular}  }
\end{table}

\paragraph{Logic.}
We evaluate DGRO in two settings: \textit{in-distribution} evaluation on the training set and \textit{out-of-distribution} (OOD) evaluation on held-out data.
We use 4.5k examples containing between 3 to 7 participants to train DGRO, and the results for $2$ and $8$ participants is the OOD prediction results. Training is performed with a batch size of $4$, using $8$ rollouts per prompt. The base model DGRO trains on is \texttt{Qwen2.5-7B-Instruct-1M}. Table \ref{tab:eval_logic} records the pass$@1$ results for the base models like \texttt{Deepseek-Math-7B}, \texttt{Qwen2.5-7B-Instruct-1M}, results for reasoning model Logic-RL \cite{xie2025logic}, our DGRO, and results for some closed-source commercial models. 
Results displayed here are under a fixed $(\beta_1, \beta_2) = (1, 0.1)$ setting, and the reward ranges we choose are $r\in [-3, 3]$ and $r\in [-5, 5]$.
Bold values indicate optimal results, and underlined values indicate suboptimal results.
In the in-distribution test results, DGRO achieves the state-of-the-art performance in almost all cases. More importantly, in the out-of-distribution setting, DGRO demonstrates strong generalization to harder tasks. 
Nevertheless, DGRO achieves an accuracy of 87\% on this setting, outperforming both DeepSeek-R1 and O3-mini-high. This demonstrates  DGRO’s effectiveness in enhancing the LLM’s generalization capabilities on complex reasoning tasks.

\paragraph{Math.}
For the mathematical reasoning tasks, we train DGRO model respectively on the DeepScaleR-Preview-Dataset (deepscaler) \cite{deepscaler2025} and Open-Reasoner-Zero (orz) \cite{hu2025orz}, and evaluate each model on a common set of benchmarks:
AIME 2024, MATH 500, AMC 2023, Minerva Math, and Olympiad Bench. 
The DeepScaleR-Preview-Dataset comprises AIME (1984–2023), AMC (excluding 2023), Omni-MATH \cite{gao2024omni}, and Still \cite{min2024imitate} datasets, totaling approximately 40.3k samples. The orz dataset includes AIME (up to 2023), MATH \cite{hendrycksmath2021},
 Numina-Math collection \cite{numina_math_datasets}, Tulu3 MATH \cite{lambert2024t}, OpenR1-Math-220k, and other open source datasets.
 
We adopt \texttt{DeepSeek-R1-Distill-Qwen-7B (8k)} (trained with the maximum response length of 8k) as the base model for RL training on both two datasets. 
The maximum response length is set to $8,192$ tokens, the temperature parameter is $1$, and $(\beta_1, \beta_2) = (1, 0.1)$.
In Table \ref{tab:math-eval}, our DGRO significantly improves the reasoning ability compared to the base model.


    

\begin{table}[ht]
\centering
\caption{Evaluation of reasoning and base models on the Math datasets with pass$@1$ as the metric.}
\renewcommand{\arraystretch}{1.2}
\resizebox{0.98\linewidth}{!}{
\begin{tabular}{lccccc|c}
\toprule
\textbf{Model} & \makecell{\textbf{AIME} \\ \textbf{2024}} & \makecell{\textbf{MATH} \\ \textbf{500}} & \makecell{\textbf{AMC} \\ \textbf{2023}} & \makecell{\textbf{Minerva} \\ \textbf{Math}} & \makecell{\textbf{Olympiad} \\ \textbf{Bench}} & \textbf{Avg.} \\
\midrule
O1-Preview & 0.40 & 0.81 & - & - & - & - \\
GPT-4o-2024-08-06 & - & 0.81 & - & 0.37 & 0.43 & - \\
Deepseek-R1 & 0.80 & 0.97 & - & - & - & - \\
\midrule
Qwen2.5-Math-7B-Instruct & 0.13 & 0.50 & 0.51 & 0.35 & 0.41 & 0.44 \\
rStar-Math-7B & 0.27 & 0.78 & 0.48 & - & 0.47 & - \\
Eurus-2-7B-PRIME & 0.28 & 0.58 & 0.39 & - & 0.42 & 0.42 \\
Qwen2.5-7B-SimpleRL & 0.30 & 0.63 & 0.40 & 0.43 & 0.43 & 0.51 \\
DeepScalek-1.5B-Preview & 0.43 & 0.88 & 0.74 & 0.30 & 0.50 & 0.57 \\
DeepSeek-R1-Distill-Qwen-1.5B & 0.29 & 0.83 & 0.27 & - & 0.43 & 0.49 \\
DeepSeek-R1-Distill-Qwen-7B (32k) & 0.56 & 0.93 & - & - & - & - \\
DeepSeek-R1-Distill-Qwen-7B (8k) & 0.40 & 0.86 & 0.69 & 0.35 & 0.49 & 0.558 \\
\rowcolor{bg!70}\textbf{+DGRO}, Distill-Qwen-7B (8k) on Deepscaler & \textbf{0.43} & \textbf{0.91} & \textbf{0.81} & \textbf{0.39} & \textbf{0.53} & \textbf{0.614} \\
\rowcolor{bg!70}\textbf{+DGRO}, Distill-Qwen-7B (8k) on Orz & \textbf{0.50} & \textbf{0.92} & \textbf{0.81} & 0.35 & \textbf{0.56} & \textbf{0.628}  \\
\bottomrule
\end{tabular}
}
\label{tab:math-eval}
\end{table}

\subsection{Ablation Study}
\label{sec:ablation}
As shown in Theorem~\ref{theorem:var}, both decoupled $\beta_1$, $\beta_2$, and the variance of the reward values influence the gradient magnitude during training. To empirically validate the trade-off of exploration and exploitation, we conduct ablation experiments focusing on these factors on K\&K logic dataset.

\begin{figure}[htbp]
    \centering
     \begin{subfigure}[t]{0.325\textwidth}
        \centering
    \includegraphics[width=\textwidth]{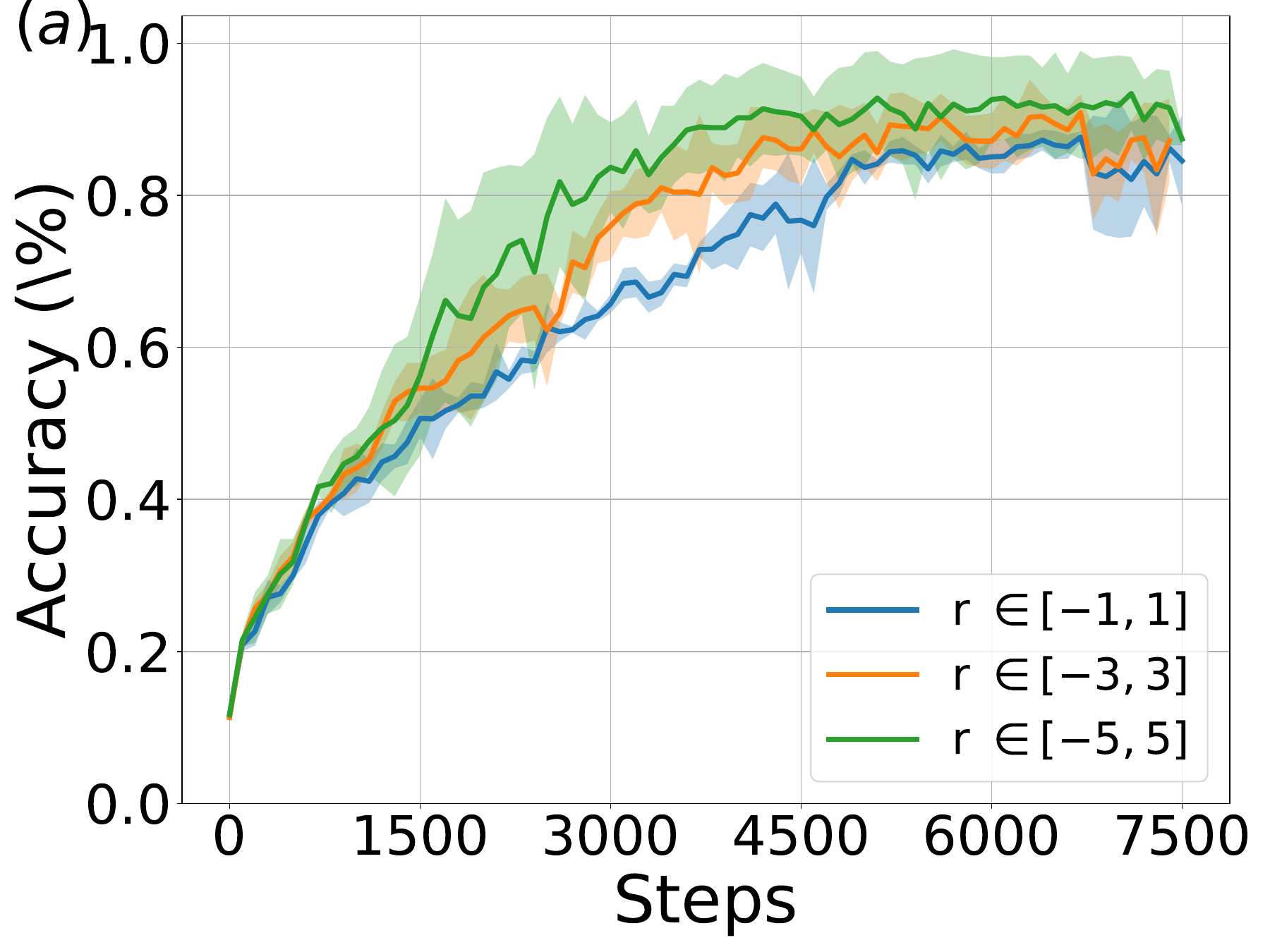}
    \end{subfigure}
    \hfill
    \begin{subfigure}[t]{0.325\textwidth}
        \centering
        \includegraphics[width=\textwidth]{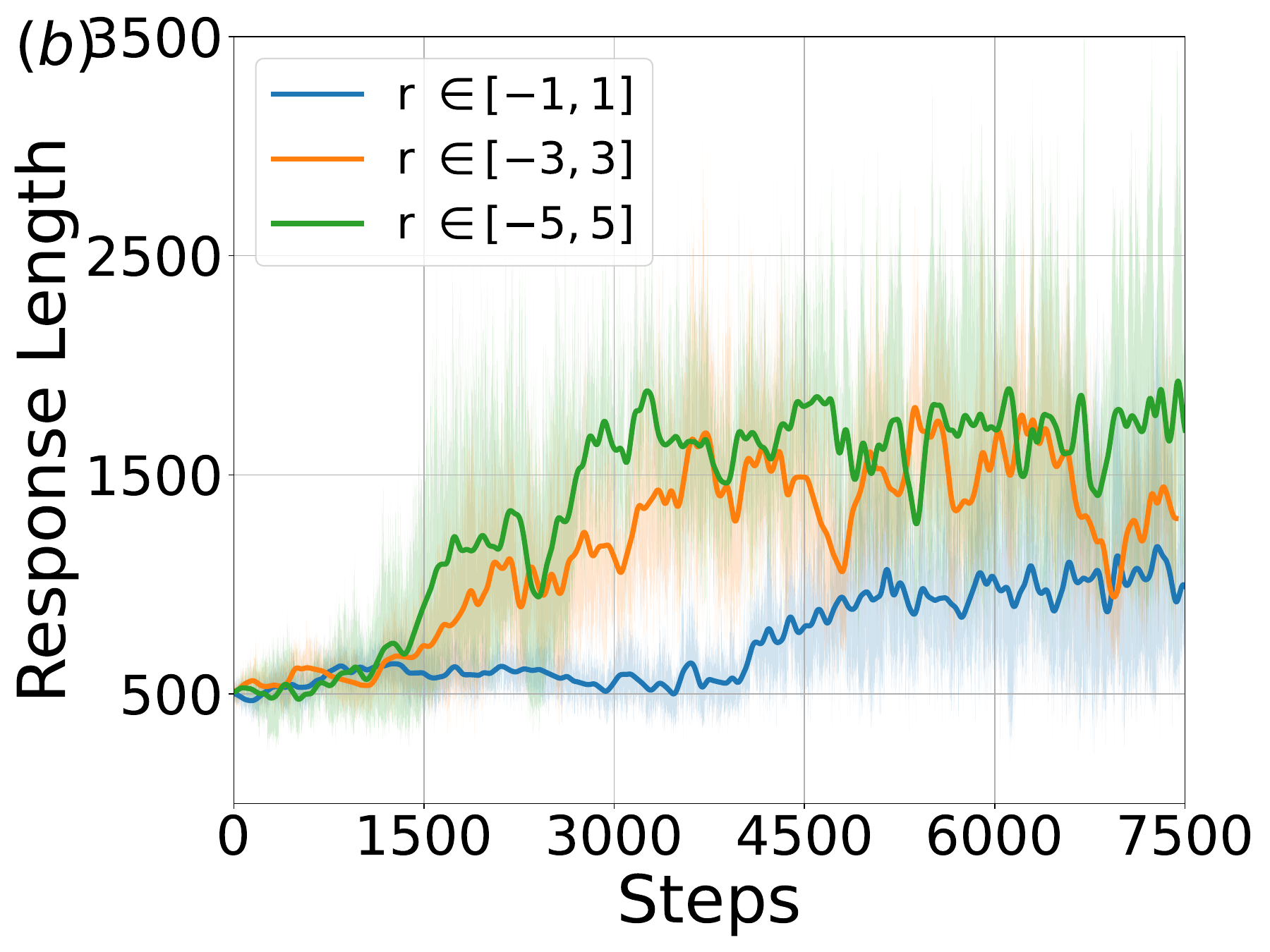}
    \end{subfigure}
    \hfill
    \begin{subfigure}[t]{0.325\textwidth}
        \centering
        \includegraphics[width=\textwidth]{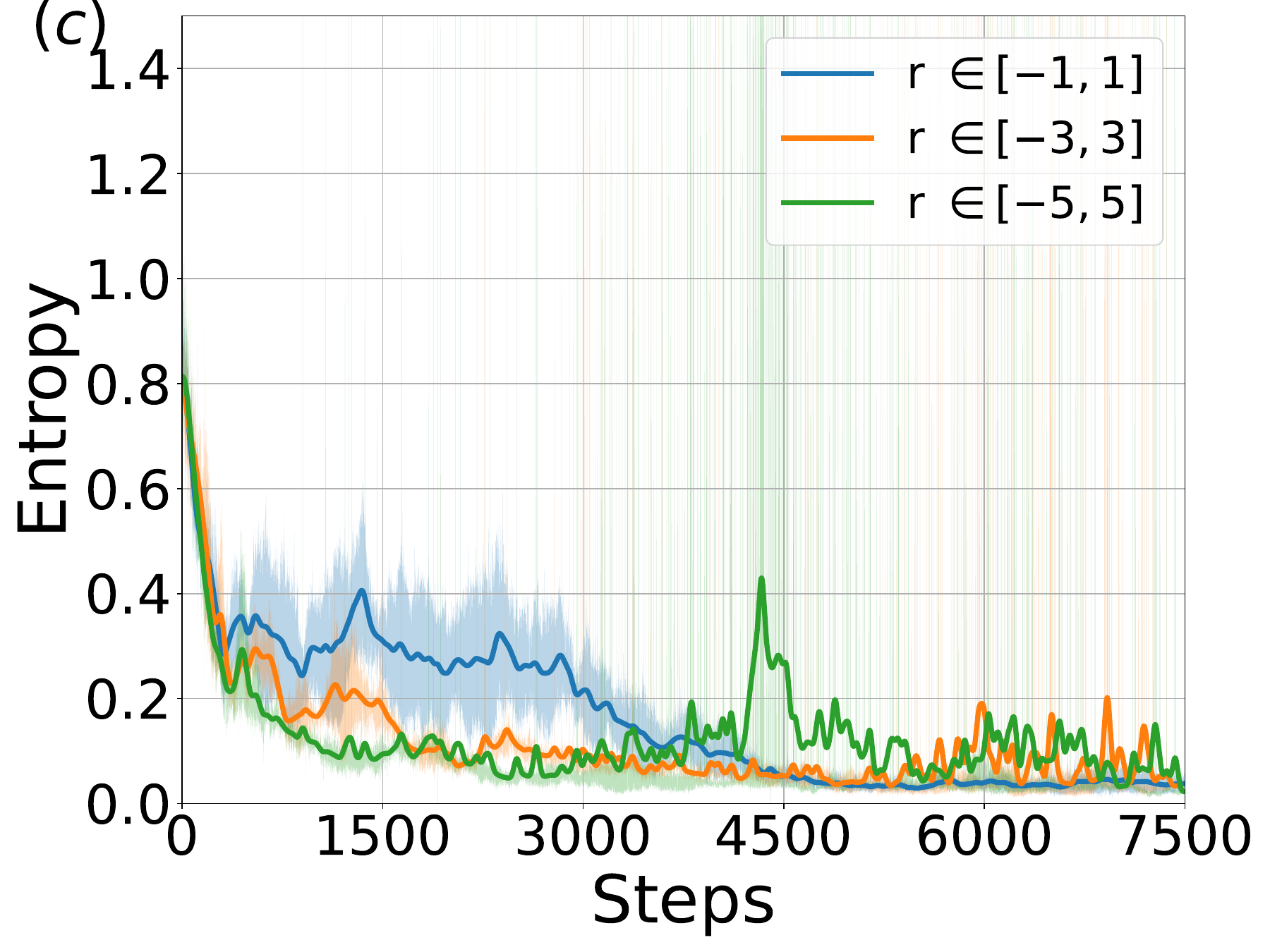}
    \end{subfigure}
    \caption{Comparison of (a) accuracy, (b) response length, and (c) entropy for different reward ranges on the K\&K dataset using DGRO. In (a), solid lines denote mean accuracy over three seeds, with shaded areas indicating variance. In (b) and (c), solid lines represent smoothed means, and shaded areas show raw measurements.}
    \label{fig:vary_reward}
    \vspace{-0.05in}
\end{figure}

\paragraph{Reward variance.}
To examine the effect of reward variance, we manipulate the reward range to control the variance explicitly. Specifically, we evaluate three reward settings with ranges $[-1, 1]$, $[-3, 3]$, and $[-5, 5]$, yielding empirical variances of $0.5075$, $4.6719$, and $12.6875$, respectively. The specific setting can be seen in Table \ref{table:reward_design} in Appendix \ref{sec:more_exp}. All experiments are conducted with a fixed temperature $T = 1.0$. For each setting, we train the model using three random seeds and report the mean and standard deviation across runs (see Figure~\ref{fig:vary_reward}).

Figure~\ref{fig:vary_reward} presents three subfigures: (a) accuracy, (b) the length of generated outputs, and (c) the entropy. We fix the coefficients $\beta_1 = 1$ and $\beta_2 = 0.1$ throughout. The shadows in Figure~\ref{fig:vary_reward} (a) is data variance for three accuracy results, and the solid lines are the results evaluated. Shadows in Figure~\ref{fig:vary_reward} (b) and Figure~\ref{fig:vary_reward} (c) are original data, and solid lines are the smoothed curve using a 1D Gaussian filter with a standard deviation of $\sigma = 25$.

As shown in Figure~\ref{fig:vary_reward} (a), increasing reward variance generally accelerates learning, evidenced by sharper accuracy gains—aligning with the theoretical prediction in Theorem~\ref{theorem:var}. In Figure~\ref{fig:vary_reward} (b), we observe that accuracy improvements under DGRO do not lead to excessively long reasoning chains, indicating that the method maintains training efficiency. On the relative simple Logic dataset, all reward settings eventually converge to low-entropy solutions. The high-variance setting occasionally shows entropy spikes, suggesting enhanced exploration, which may contribute to its superior performance.
In contrast, Math presents a more complex challenge. As shown in Figure~\ref{fig:math} (Appendix~\ref{sec:more_exp}), the entropy curve rises gradually throughout training, indicating that math requires continuous exploration by DGRO.

\paragraph{Influence of decoupled $\beta_1$ and $\beta_2$.}
$\beta_1$ in DGRO scales the policy gradient, thereby influencing the degree of exploration, while $\beta_2$ controls the strength of KL regularization between the current policy and the policy last round, which influences the exploitation.
To study their effects, we evaluate DGRO under various $(\beta_1, \beta_2)$ configurations.

\begin{figure}[htbp]
    \centering
    \begin{subfigure}[t]{0.32\textwidth}
        \centering
        \includegraphics[width=\textwidth]{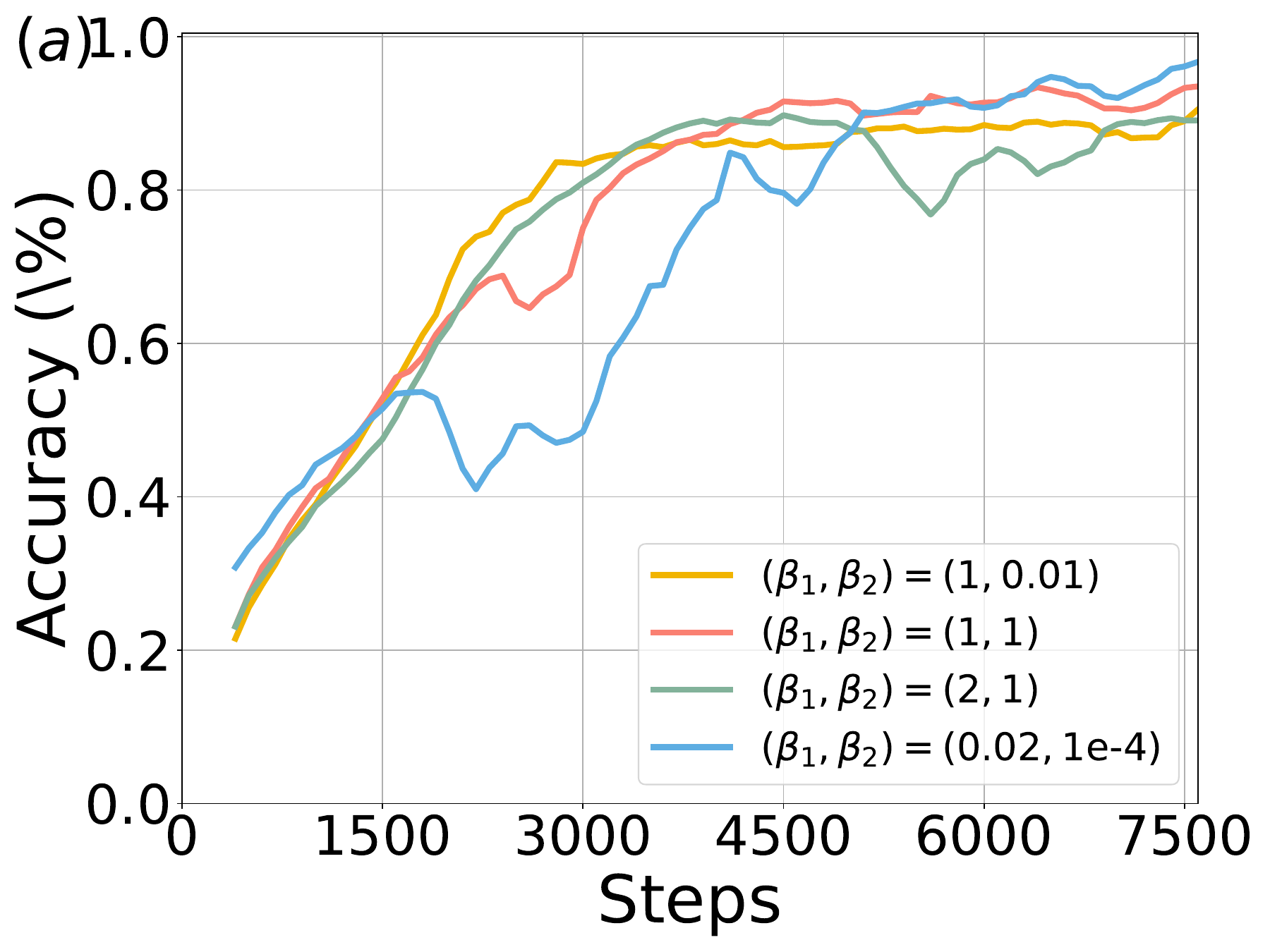}
    \end{subfigure}
    \hfill
    \begin{subfigure}[t]{0.32\textwidth}
        \centering
        \includegraphics[width=\textwidth]{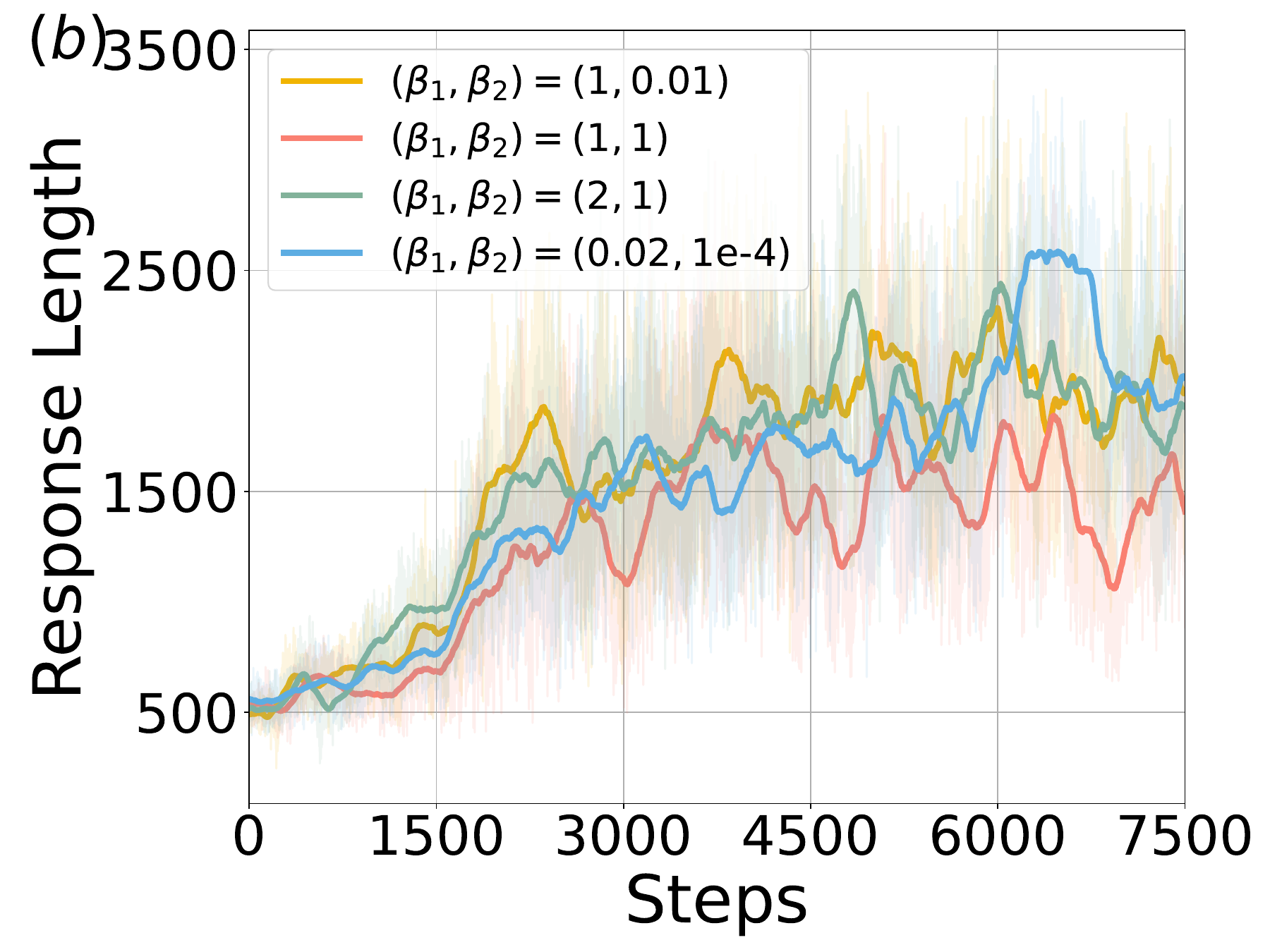}
    \end{subfigure}
    \hfill
    \begin{subfigure}[t]{0.32\textwidth}
        \centering
        \includegraphics[width=\textwidth]{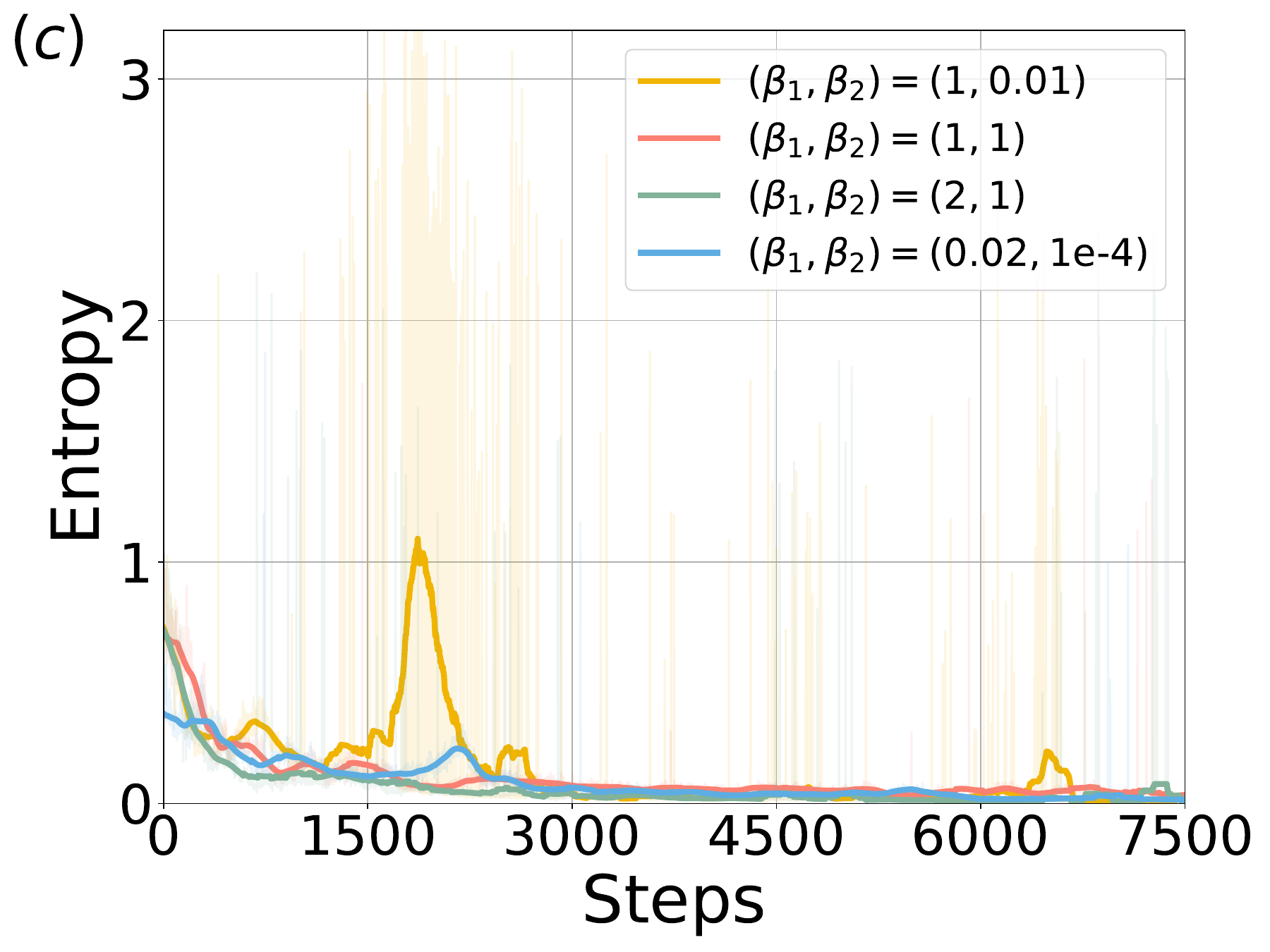}
    \end{subfigure}
    \caption{Comparison of (a) accuracy, (b) response length, and (c) entropy during training for different $(\beta_1, \beta_2)$ values.}
    \label{fig:vary_beta}
    \vspace{-0.05in}
\end{figure}
Figure~\ref{fig:vary_beta} visualizes the learning dynamics. Figure~\ref{fig:vary_beta} (a) presents evaluation accuracy, smoothed using a moving average with a window size of $5$. 
When $\beta_1$ is not very small, reducing $\beta_2$, the accuracy curve becomes steeper, indicating less KL constrains and faster policy improvement. Consistent with Theorem \ref{theorem:error_of_taulogz-small_tau}, when $\beta_2$ is relatively small, the approximation error of the reward will decrease and the reasoning ability may increase.
Furthermore, as shown in the entropy Figure~\ref{fig:vary_beta} (c), the configuration with $\beta_1 = 1$ and $\beta_2 = 0.01$ exhibits noticeable fluctuations in entropy, suggesting that a smaller $\beta_2$ facilitates more exploration during the reasoning process.

\begin{figure}[htbp]
    \centering
    \begin{subfigure}[t]{0.98\textwidth}
        \centering
        \includegraphics[width=\textwidth]{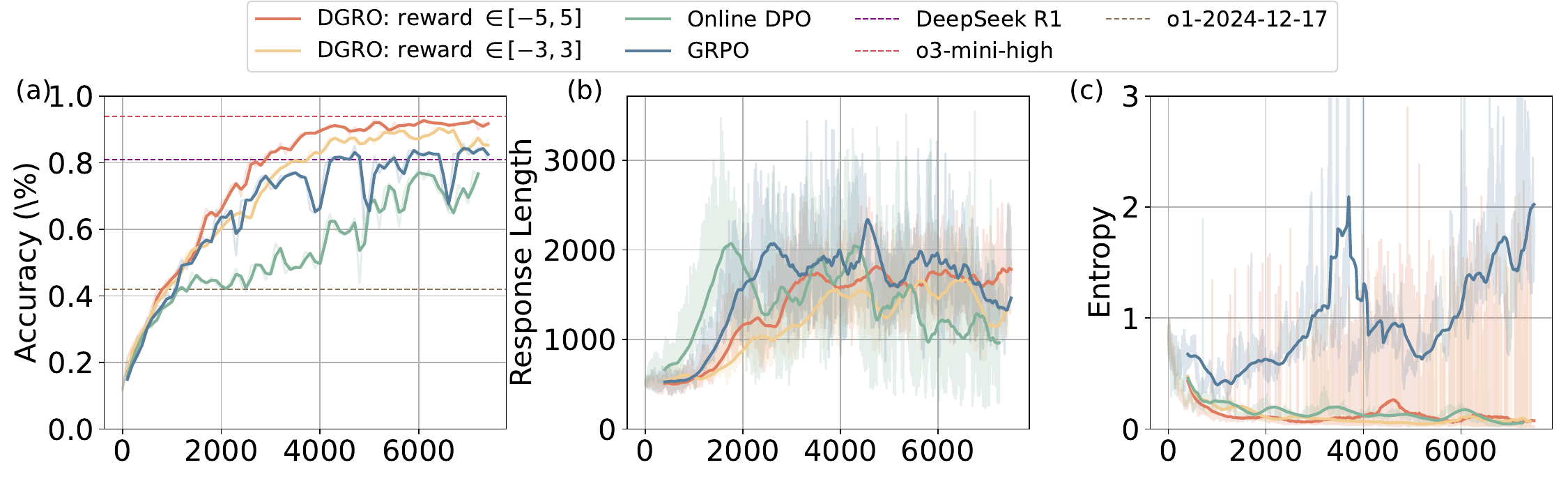}
    \end{subfigure}
    \caption{Comparison of (a) accuracy, (b) response length, and (c) entropy for different models: DGRO with reward $\in [-5, 5]$ (red), DGRO with reward $\in [-3, 3]$ (yellow), Online DPO (green) and GRPO (blue). DGRO curves show mean performance over three seeds.}
    \vspace{-0.1in}
    \label{fig:compare_models}
\end{figure}

\paragraph{Comparisons of exiting methods.} 
To further evaluate DGRO, we compare it with existing reasoning algorithms in terms of stability, accuracy, and efficiency \cite{DeepSeekR1, qi2024online}. We present the performance of DGRO, GRPO \cite{shao2024deepseekmath}, and Online DPO \cite{guo2024direct} across three metrics in Figure~\ref{fig:compare_models}: (a) accuracy, (b) response length, and (c) entropy.

In terms of accuracy, to reduce the impact of randomness, each DGRO setting is conducted three times with different random seeds. The average accuracy curves across these runs are shown in Figure~\ref{fig:compare_models} (a). The results demonstrate that DGRO consistently outperforms GRPO and Online DPO under both the $[-3, 3]$ and $[-5, 5]$ reward ranges. It ultimately converges to a level that is on par with that of \texttt{o3-mini-high} (pink dashed line). In contrast, GRPO performs worse and converges to the level of DeepSeek R1 (purple dashed line). 
Notably, DGRO remains efficient even in high-accuracy settings, without extra training cost. As shown in Figure \ref{fig:compare_models} (b), DGRO does not lead to increased response length during training, demonstrating its efficiency.
Therefore, DGRO demonstrates superior reasoning capabilities across multiple aspects.

\section{Conclusions, Discussions and Limitations}
\label{sec:conclusion}
In this paper, we introduce Decoupled Group Reward Optimization (DGRO), a novel online reinforcement learning algorithm, which decouples the regularization coefficient in standard RL objectives into the policy gradient coefficient, and the policy distance coefficient. Through different configuration choices, DGRO can recover other reward optimization algorithm variants. Our analysis reveals that this decoupling strategy offers a better balance between exploration and exploitation.
Theoretically, we show that the regularization coefficients significantly affect the convergence behavior. Empirically, we conduct ablation studies on the K\&K dataset to validate these findings. Moreover, we analyze the validity of using the reward mean to approximate the soft value function $V(x, \beta)$ under various conditions. The analysis offers theoretical support and practical guidance for the design and optimization of the DGRO and other related works. 
Experimental results show that DGRO achieves competitive performance on the K\&K and math datasets and consistently outperforms the base model.

\paragraph{Discussions and Limitations.} DGRO enables a clearer analysis of optimization dynamics by decoupling key parameters that govern exploration and exploitation. This decoupling not only improves flexibility but also allows individual assessment of each component's influence on reasoning performance. Our study further reveals the impact of reward variance, highlighting its non-negligible role in model behavior. However, potential interactions between parameters, as well as other latent factors affecting reasoning, remain underexplored. We leave these aspects for future investigation.
\bibliographystyle{unsrt}
\bibliography{references}

\appendix

\section{Discussion of DGRO and other special case}\label{appendix:DGRO_and_others}

In this section, we discuss the relationship between DGRO and prior works such as Direct Reward Optimization (DRO) \cite{DRO_Deepmind} and Kimi k1.5 \cite{team2025kimi}. Both DRO and Kimi k1.5 can be viewed as specific instances within the broader DGRO framework, distinguished by particular configurations of hyperparameters and design choices, as shown in Table \ref{tab:DGRO_and_others}. For example, Kimi k1.5 employs a fixed scalar $\beta$ and approximates $V(x, \beta)$ using the empirical mean $\bar r(x)$ of sampled rewards, while DRO utilizes a learned value function $V_\phi(x)$ to estimate $V(x, \beta)$ and adopts different regularization strategies. By adjusting components such as the policy gradient coefficient $\beta_1$, the regularization coefficient $\beta_2$, the choice of value function $V(x, \beta)$, and the reference policy $\pi_{\theta_{\text{src}}}$, DGRO provides a flexible and unified framework that encapsulates these methods. This generalization facilitates a comprehensive understanding of various reward optimization strategies and their interconnections.


To avoid introducing additional technical complexities into our analysis, we clarify that the comparisons and discussions presented here are limited strictly to the perspective of RL algorithm loss functions. The derivation from DGRO to DRO \cite{DRO_Deepmind} is given below.
\begin{equation}
\small{  \begin{aligned}
\mathcal{L}_{\text{DGRO}}(\theta) &= -\mathbb{E}_{x \sim \mathcal{D},y \sim \pi_{{\text{sam}}}(\cdot|x)} \left[\beta_1
\left( r(x, y) - V(x,\beta) \right)\log \pi_\theta(y \mid x) 
- \beta_2\left(\log \frac{\pi_\theta\left(y \mid x\right)}{\pi_{\text{src}}\left(y \mid x\right)}\right)^2
\right]\\
&\overset{(*)}{=}
 \mathbb{E}_{x \sim \mathcal{D},y \sim \mu(\cdot|x)} \left[-\beta
\left( r(x, y) - V_\phi(x) \right)\log \pi_\theta(y \mid x) 
+ \frac{\beta^2}{2}\left(\log \frac{\pi_\theta\left(y \mid x\right)}{\pi_{\text{ref}}\left(y \mid x\right)}\right)^2
\right]\\
&\equiv \frac{1}{2}\mathbb{E}_{x \sim \mathcal{D},y \sim \mu(\cdot|x)} \left[\left(r(x, y) - V_\phi(x) - \beta\log \frac{\pi_\theta\left(y \mid x\right)}{\pi_{\text{ref}}\left(y \mid x\right)}\right)^2
\right]\\
&=\mathcal{L}_{\text{DRO}}(\theta),\\
\text{(*) denotes }&\text{the adoption of the parameter settings associated with DRO.}
\end{aligned}}
\end{equation}
where ``$\equiv$" means that the left and right ends differ by a term that is unrelated to $\theta$. The right side of the last equation corresponds to the DRO objective of Eq.4 in Ref. \cite{DRO_Deepmind}.

The derivation from DGRO to Kimi k1.5 \cite{team2025kimi} is given below.
\begin{equation}
{  \begin{aligned}
\mathcal{L}_{\text{DGRO}}(\theta) &= -\mathbb{E}_{x \sim \mathcal{D},y \sim \pi_{{\text{sam}}}(\cdot|x)} \left[\beta_1
\left( r(x, y) - V(x,\beta) \right)\log \pi_\theta(y \mid x) 
- \beta_2\left(\log \frac{\pi_\theta\left(y \mid x\right)}{\pi_{\text{src}}\left(y \mid x\right)}\right)^2
\right]\\
&\overset{(**)}{=} \mathbb{E}_{x \sim \mathcal{D},y \sim \pi_{\theta_{\text{old}}}(\cdot|x)} \left[-
\left( r(x, y) - V_\phi(x) \right)\log \pi_\theta(y \mid x) 
+ \frac{\beta}{2}\left(\log \frac{\pi_\theta\left(y \mid x\right)}{\pi_{\theta_{\text{old}}}\left(y \mid x\right)}\right)^2
\right]\\
&\equiv \frac{1}{2\beta}\mathbb{E}_{x \sim \mathcal{D},y \sim \pi_{\theta_{\text{old}}}(\cdot|x)} \left[\left(r(x, y) - V_\phi(x) - \beta\log \frac{\pi_\theta\left(y \mid x\right)}{\pi_{\theta_{\text{old}}}\left(y \mid x\right)}\right)^2
\right]\\
&=\mathcal{L}_{\text{Kimi k1.5}}(\theta),\\
\text{(**) denotes }&\text{the adoption of the parameter settings associated with Kimi k1.5.}
\end{aligned}}
\end{equation}
where ``$\equiv$" means that the left and right ends differ by a term that is unrelated to $\theta$. The right side of the second-to-last equation corresponds to the DRO objective of Eq.3 in Ref. \cite{team2025kimi}.

Similarly, we rewrite DGRO into the MSE format.
\begin{equation}\label{eq:DGRO_beta_corresponding}
{  \begin{aligned}
\mathcal{L}_{\text{DGRO}}(\theta) &= -\mathbb{E}_{x \sim \mathcal{D},y \sim \pi_{{\text{sam}}}(\cdot|x)} \left[\beta_1
\left( r(x, y) - V(x,\beta) \right)\log \pi_\theta(y \mid x) 
- \beta_2\left(\log \frac{\pi_\theta\left(y \mid x\right)}{\pi_{\text{src}}\left(y \mid x\right)}\right)^2
\right]\\
&\overset{(***)}{=} \mathbb{E}_{x \sim \mathcal{D},y \sim \pi_{\theta_{\text{old}}}(\cdot|x)} \left[-\beta_1
\left( r(x, y) - \bar r(x) \right)\log \pi_\theta(y \mid x) 
+ \beta_2\left(\log \frac{\pi_\theta\left(y \mid x\right)}{\pi_{\theta_{\text{old}}}\left(y \mid x\right)}\right)^2
\right]\\
&\equiv \frac{\beta_1^2}{4\beta_2}\mathbb{E}_{x \sim \mathcal{D},y \sim \pi_{\theta_{\text{old}}}(\cdot|x)} \left[\left(r(x, y) - \bar r(x) - \frac{2\beta_2}{\beta_1}\log \frac{\pi_\theta\left(y \mid x\right)}{\pi_{\theta_{\text{old}}}\left(y \mid x\right)}\right)^2
\right],\\
\text{(***) }&\text{denotes the adoption of the parameter settings associated with the implementations of DGRO.}
\end{aligned}}
\end{equation}
where ``$\equiv$" means that the left and right ends differ by a term that is unrelated to $\theta$. According to Eq.\ref{eq:DGRO_beta_corresponding}, the corresponding relationship between the original regularization coefficient $\beta$ and the decoupled coefficients $(\beta_1,\beta_2)$ is $\beta=\frac{2\beta_2}{\beta_1}$.

\section{Proof of Lemma \ref{lemma:unbiased_estimation}}\label{proof:unbiased_estimation}

Recall that $V(x, \beta) = \beta \log Z(x)$, where the partition function is defined as $Z(x) = \sum_y \pi_{\text{src}}(y \mid x) \exp \left(\frac{1}{\beta} r(x, y)\right)$. We reformulate Lemma \ref{lemma:unbiased_estimation} in the following form.

\textbf{Lemma \ref{lemma:unbiased_estimation}:} Denote $\bar{r}(x)=\mathbb{E}_{y\sim \pi_{{\text{sam}}}(\cdot|x)}[r(x,y)]$, $V(x, \beta)=\beta\log\sum_y \pi_{\text{src}}(y \mid x) \exp \left(\frac{1}{\beta} r(x, y)\right)$. When $\pi_{{\text{sam}}}=\pi_{{\text{src}}}$, the following equation holds if and only if $\beta\to\infty$ or $r(x,y)=c(x)$ holds almost everywhere under measure $\pi_{{\text{sam}}}(\cdot|x)$, where $c(x)$ depends only on $x$:
\begin{equation}
    \mathbb{E}_{y\sim \pi_{{\text{sam}}}(\cdot|x)}[r(x,y)] = \beta\log\left(\mathbb{E}_{y\sim \pi_{{\text{src}}}(\cdot|x)}\left[\exp\left(\frac{r(x,y)}{\beta}\right)\right]\right).
\end{equation}

\textbf{Proof:} For simplicity, we denote $\pi_{{\text{sam}}}(\cdot|\cdot)=\pi_{{\text{src}}}(\cdot|\cdot)=p(\cdot|\cdot)$. Firstly we prove the situation of $r(x,y)=c(x)$. Denote $A = \mathbb{E}_{y\sim p(\cdot|x)}[r(x,y)]$ and $ B = \mathbb{E}_{y\sim p(\cdot|x)}\left[\exp\left(\frac{r(x,y)}{\beta}\right)\right]$. Consider when the formula $\exp(A) - B^\beta$ equals to zero, i.e.
\begin{equation}
    \exp(A) - B^\beta=0\quad \text{or}\quad A = \beta\ln B.
\end{equation}

Since $f(z)=\exp(z)$ is a convex function, according to Jensen's inequality, we can get 
\begin{equation}
    \exp\big(\mathbb{E}_{y\sim p(\cdot|x)}[Z]\big) \leq \mathbb{E}_{y\sim p(\cdot|x)}\big[\exp(Z)\big],
\end{equation}
which takes the same sign if and only if and only if \(Z\) takes the same constant almost everywhere under the measure \(p(\cdot|x)\).

Let $Z = \frac{r(x,y)}{\beta}$, then
\begin{equation}
\exp\left(\mathbb{E}_{y\sim p(\cdot|x)}\left[\frac{r(x,y)}{\beta}\right]\right) \leq \mathbb{E}_{y\sim p(\cdot|x)}\left[\exp\left(\frac{r(x,y)}{\beta}\right)\right]=B.
\end{equation}

Taking the logarithm and multiplying by \(\beta\), then we have:
\begin{equation}
    \mathbb{E}_{y\sim p(\cdot|x)}[r(x,y)] \leq \beta\ln B.
\end{equation}

The lower bound of $\beta\ln B$ given by Jensen’s inequality is attained which means the equality \(\mathbb{E}_{y\sim p(\cdot|x)}[r(x, y)] = \beta \ln B\) holds if and only if  
\begin{equation}
\frac{r(x, y)}{\beta} \quad \text{(equivalently, \(r(x, y)\)) is constant almost everywhere}.
\end{equation}   
That is, \(r(x, y)\) takes the same constant value for almost all \(y\) under the support of \(p(\cdot \mid x)\). Denoting this constant by \(c\), we then have  
\begin{equation}
\mathbb{E}_{y\sim p(\cdot|x)}[r(x, y)] = c(x) \quad \text{and} \quad \mathbb{E}_{y\sim p(\cdot|x)}\left[\exp\left(\frac{r(x, y)}{\beta}\right)\right] = \exp\left(\frac{c(x)}{\beta}\right).
\end{equation}   
Substituting these into the expression of A and B yields  
\begin{equation}
\exp(A) - B^\beta=\exp(c(x)) - \left(\exp\left(\frac{c(x)}{\beta}\right)\right)^\beta = \exp(c(x)) - \exp(c(x)) = 0.
\end{equation} 

Secondly, we prove the situation of $\beta\to\infty$. Consider the expression:
\begin{equation}
    \exp\left(\mathbb{E}_{y \sim  p(\cdot \mid x)}[r(x, y)]\right) - \left(\mathbb{E}_{y \sim  p(\cdot \mid x)}\left[\exp\left(\frac{r(x, y)}{\beta}\right)\right]\right)^{\beta}.
\end{equation} 

\begin{equation}
     \left(\mathbb{E}_{y \sim  p(\cdot \mid x)}\left[\exp\left(\frac{r(x, y)}{\beta}\right)\right]\right)^{\beta} - \exp\left(\mathbb{E}_{y \sim  p(\cdot \mid x)}[r(x, y)]\right).
\end{equation}

For each \(r(x, y)\), when \(\beta\) is large, the term \(\frac{r(x, y)}{\beta}\) becomes small, apply a Taylor expansion, we have:
\begin{equation}
\exp\left(\frac{r(x, y)}{\beta}\right) = 1 + \frac{r(x, y)}{\beta} + \frac{r(x, y)^2}{2\beta^2} + O\left(\frac{1}{\beta^3}\right).
\end{equation} 

Taking the expectation with respect to \(y \sim  p(\cdot \mid x)\), we obtain:
\begin{equation}
\mathbb{E}\left[\exp\left(\frac{r(x, y)}{\beta}\right)\right] = \mathbb{E}\left[1 + \frac{r(x, y)}{\beta} + \frac{r(x, y)^2}{2\beta^2} + O\left(\frac{1}{\beta^3}\right)\right].
\end{equation} 
Since expectation is linear, the following equation yields:
\begin{equation}
\mathbb{E}\left[\exp\left(\frac{r(x, y)}{\beta}\right)\right] = 1 + \frac{\mathbb{E}[r(x, y)]}{\beta} + \frac{\mathbb{E}[r(x, y)^2]}{2\beta^2} + O\left(\frac{1}{\beta^3}\right).
\end{equation} 

Define $A_\beta = \mathbb{E}\left[\exp\left(\frac{r(x, y)}{\beta}\right)\right]$, so that
\begin{equation}
A_\beta \approx 1 + \frac{\mathbb{E}[r(x, y)]}{\beta} + O\left(\frac{1}{\beta^2}\right).
\end{equation} 

For any constant \(a\), we have the classical limit:
$\left(1 + \frac{a}{\beta}\right)^\beta \to \exp(a)$ as $\beta \to \infty$. 

In our case, setting \(a = \mathbb{E}[r(x, y)]\), it follows that
\begin{equation}
\left(1 + \frac{\mathbb{E}[r(x, y)]}{\beta}\right)^\beta \to \exp(\mathbb{E}[r(x, y)]).
\end{equation} 

Since \(A_\beta \approx 1 + \frac{\mathbb{E}[r(x, y)]}{\beta}\), and the error term \(O(1/\beta^2)\) vanishes in the limit, we conclude that
\begin{equation}
A_\beta \to \exp(\mathbb{E}[r(x, y)]) \quad \text{as } \beta \to \infty.
\end{equation}

Substituting the results above into the original expression yields:
\begin{equation}
\lim_{\beta \to \infty} \left\{ \exp(\mathbb{E}[r(x, y)]) - \left(\mathbb{E}\left[\exp\left(\frac{r(x, y)}{\beta}\right)\right]\right)^{\beta} \right\} = \exp(\mathbb{E}[r(x, y)]) - \exp(\mathbb{E}[r(x, y)]) = 0.
\end{equation} 

Proof finished. $\hfill\qedsymbol$ 

\section{Proof of Lemma \ref{lemma:biased_estimation}}\label{proof:biased_estimation}
Recall that $V(x, \beta) = \beta \log Z(x)$, where the partition function is defined as $Z(x) = \sum_y \pi_{\text{src}}(y \mid x) \exp \left(\frac{1}{\beta} r(x, y)\right)$. We reformulate Lemma \ref{lemma:unbiased_estimation} in the following form.

\textbf{Lemma \ref{lemma:biased_estimation}:} Denote $m(x)=\sup _y r(x, y)$. Assume $\pi_{{\text{sam}}}=\pi_{{\text{src}}}$. In the limit $\beta \to 0$, a non-zero error exists between $\bar{r}(x)$ and $\beta \log\left(Z(x)\right)$, if the reward function is not constant:
\begin{equation}
\lim_{\beta\to0}\beta\log\left(Z(x)\right) - \bar{r}(x) = m(x)-\bar{r}(x).
\end{equation}

\textbf{Proof:} For simplicity, we denote $\pi_{{\text{sam}}}(\cdot|\cdot)=\pi_{{\text{src}}}(\cdot|\cdot)=p(\cdot|\cdot)$. The proof of Lemma \ref{lemma:biased_estimation} is straightforward. The key idea is that as $\beta\to0$, the “soft‐max” (log‐sum‐exp) operator concentrates on the largest reward. Specifically,
\begin{equation}
\lim_{\beta\to0}\beta\log\left(\mathbb{E}_{y\sim p(\cdot\mid x)}\bigl[\exp\bigl(\tfrac{1}{\beta}r(x,y)\bigr)\bigr]\right)=\max_{y}r(x,y).
\end{equation}

Denote $\delta(y)=m(x) - r(x,y)\ge0$. Then
\begin{equation}
    r(x,y) = M - \delta(y)
     \quad\Longrightarrow\quad
     \exp\bigl(r(x,y)/\beta\bigr)
     = e^{M/\beta}e^{-\delta(y)/\beta}.
\end{equation}

Hence $\mathbb{E}_{y\sim p(\cdot\mid x)}\bigl[e^{r(x,y)/\beta}\bigr]= e^{M/\beta}\mathbb{E}_{y\sim p(\cdot\mid x)}\bigl[e^{-\delta(y)/\beta}\bigr]$. Taking logarithms and multiplying by $\beta$ gives
\begin{equation}
\beta\log\left(\mathbb{E}_{y\sim p(\cdot\mid x)}\left[e^{r(x,y)/\beta}\right]\right)= \beta\left(\frac{M}{\beta} +\log\mathbb{E}_{y\sim p(\cdot\mid x)}\left[e^{-\delta(y)/\beta}\right]\right)
= M+\beta\log\mathbb{E}_{y\sim p(\cdot\mid x)}\left[e^{-\delta(y)/\beta}\right].
\end{equation}

For every $y$, $0\le\delta(y)\ \Longrightarrow\ 0 < e^{-\delta(y)/\beta}\le1$. Then
\begin{equation}
0 < \mathbb{E}_{y\sim p(\cdot\mid x)}[e^{-\delta(y)/\beta}]\leq1\ \Longrightarrow\ -\infty < \log\mathbb{E}_{y\sim p(\cdot\mid x)}[e^{-\delta(y)/\beta}]\le0.
\end{equation}

Moreover, because there exists at least one $y^*$ achieving $\delta(y^*)=0$, we have $\mathbb{E}_y[e^{-\delta(y)/\beta}]\ge  p(y^*|x)\cdot1,$ so this expectation is bounded below by a positive constant independent of $\beta$. It follows that
\begin{equation}
    \log\mathbb{E}_{y\sim p(\cdot\mid x)}\bigl[e^{-\delta(y)/\beta}\bigr]
     = O(1)
     \quad\Longrightarrow\quad
     \beta\log\mathbb{E}_{y\sim p(\cdot\mid x)}\bigl[e^{-\delta(y)/\beta}\bigr]
     =O(\beta).
\end{equation}

As $\beta\to0$, $O(\beta)\to0$. Then we have:
\begin{equation}
\lim_{\beta\to0}\beta\log\left(\mathbb{E}_{y\sim p(\cdot\mid x)}\bigl[\exp\bigl(\tfrac{1}{\beta}r(x,y)\bigr)\bigr]\right)=\max_{y}r(x,y).
\end{equation}
\begin{equation}
\lim_{\beta\to0}\beta\log Z(x) - \bar{r}(x) = m(x)-\bar{r}(x).
\end{equation}
Proof finished. $\hfill\qedsymbol$ 

\section{Proof of Theorem \ref{theorem:error_of_taulogz-small_tau}}\label{proof:error_of_taulogz-small_tau}
\textbf{Theorem \ref{theorem:error_of_taulogz-small_tau}:} Denote $p_*(x)=P\left(r(x, y)=\sup _y r(x, y)\right)$, $m(x)=\sup _y r(x, y)$. Assume that $\pi_{{\text{sam}}}=\pi_{{\text{src}}}$. We have:
    \begin{equation}
0\leq V(x,\beta) - \bar{r}(x) \leq m(x)-\bar r(x) + \beta\left(\log p_*(x) + \frac{1-p_*(x)}{p_*(x)}\right).
    \end{equation} 

\textbf{Proof:} Because $\pi_{{\text{sam}}}(\cdot|\cdot)=\pi_{{\text{src}}}(\cdot|\cdot)$. denote $\Delta(y)=r(x,y)-\bar r(x)$, so that $\mathbb{E}[\Delta]=0$. Then
\begin{equation}
Z(x)=\mathbb{E}\bigl[e^{r/\beta}\bigr]=e^{\bar r(x)/\beta}\mathbb{E}\bigl[e^{\Delta/\beta}\bigr],
\end{equation}
and hence
\begin{equation}
\beta\log Z(x)=\bar r(x)+\underbrace{\beta\log\mathbb{E}\bigl[e^{\Delta/\beta}\bigr]}_{=:\mathbb{E}(\beta)}.
\end{equation}

Write the expectation as a two‐point mixture:
\begin{equation}
    \mathbb{E}\bigl[e^{\Delta/\beta}\bigr]
= p_*(x)e^{(m-\bar r)/\beta}
+(1-p_*(x))
\mathbb{E}\bigl[e^{\Delta/\beta}\bigm|r<m\bigr].
\end{equation}

Taking logarithms gives
\begin{equation}
    \mathbb{E}(\beta)
=\beta\log\Bigl(p_*e^{(m-\bar r)/\beta}
+(1-p_*)\mathbb{E}[e^{\Delta/\beta}\mid r<m]\Bigr).
\end{equation}

Factor out the dominant term $p_*e^{(m-\bar r)/\beta}$:
\begin{equation}
\begin{aligned}
\mathbb{E}(\beta)
&=\beta\Bigl(\tfrac{m-\bar r}{\beta} + \log p_*
+ \log\bigl(1 + \tfrac{1-p_*}{p_*}\mathbb{E}[e^{\Delta/\beta}\mid r<m]e^{-(m-\bar r)/\beta}\bigr)\Bigr)\\
&=(m-\bar r)+\beta(\log p_*+\log(1 + A(\beta))).
\end{aligned}
\end{equation}
where $A(\beta)=\frac{1-p_*}{p_*}\mathbb{E}\bigl[e^{\Delta/\beta}\mid r<m\bigr]e^{-(m-\bar r)/\beta}.
$

It's easy to know that $\Delta \leq m-\bar{r}$. Thus for every such $y$, $e^{\Delta/\beta}\leq^{(m-\bar{r})/\beta}$, Thus
$\mathbb{E}[e^{\Delta/\beta}\mid r<m]\le e^{(m-\bar r)/\beta}$. Therefore
\begin{equation}
A(\beta)\leq\frac{1-p_*}{p_*}e^{(m-\bar r)/\beta}e^{-(m-\bar r)/\beta}
=\frac{1-p_*}{p_*}.
\end{equation}

Since $\log(1 + x)\le x$ for all $x>-1$, we get
\begin{equation}
\beta\log(1 + A(\beta))\leq\beta A(\beta)\leq\beta\frac{1-p_*}{p_*},
\end{equation}

Putting this into the expression for $\mathbb{E}(\beta)$ yields
\begin{equation}
\begin{aligned}
\mathbb{E}(\beta)&\leq m-\bar r+\beta\left(\log p_*+\frac{1-p_*}{p_*}\right).
\end{aligned}
\end{equation}

Then
\begin{equation}
    0\leq\beta\log Z(x) - \bar r(x)
\leq m(x)-\bar r(x) + \beta\left(\log p_*+\frac{1-p_*}{p_*}\right).
\end{equation}


Proof finished. $\hfill\qedsymbol$.

\subsection{Additional observations}\label{proof:beta_coef_big_0} Define for brevity
$$
f(p)=\log p + \frac{1-p}{p}
=\frac1p-1+\log p,
\quad p\in(0,1].
$$

One checks

$$
f'(p)=\frac{p-1}{p^2}\le0,
$$

so $f$ decreases on $(0,1]$, with $f(1)=0$.  Hence for all $p\in(0,1]$,

$$
f(p)\;\ge\;0
\quad\Longrightarrow\quad
\log p + \frac{1-p}{p}\;\ge\;0.
$$




\section{Equivalence between \texorpdfstring{$\bar{r}(x)$}{r̄(x)} and \texorpdfstring{$\beta\log Z(x)$}{beta log Z(x)} in the large-\texorpdfstring{$\beta$}{beta} limit}\label{proof:error_of_taulogz}
Complementing Theorem \ref{theorem:error_of_taulogz-small_tau}, we also provide an upper bound on the error between $\bar{r}(x)$ and $\beta \log Z(x)$ in the regime where $\beta \to \infty$ with Theorem \ref{theorem:error_of_taulogz}. The upper bound in Theorem \ref{theorem:error_of_taulogz} vanishes as $\beta$ increases, establishing the asymptotic equivalence between $\bar{r}(x)$ and $\beta \log Z(x)$ in the large-$\beta$ limit.
\begin{theorem}
\label{theorem:error_of_taulogz}
Denote $\bar{r}(x)=\mathbb{E}_{y\sim \pi_{{\text{sam}}}(\cdot|x)}[r(x,y)]$, $V(x, \beta)=\beta\log\sum_y \pi_{\text{src}}(y \mid x) \exp \left(\frac{1}{\beta} r(x, y)\right)$. Assume that $\pi_{{\text{sam}}}(\cdot|\cdot)=\pi_{{\text{src}}}(\cdot|\cdot)=p(\cdot|\cdot)$, $r(x,y)$ is bounded and has finite variance. With the accuracy of the second-order truncation error of $\exp\bigl(\frac{r(x,y)}{\beta}\bigr)$, we have
    \begin{equation}
0\leq\beta\log Z(x) - \bar{r}(x)
\leq \frac{\operatorname{Var}_{y\sim p(\cdot|x)}[r(x,y)]}{2\beta}.
    \end{equation}
\end{theorem}

\textbf{Proof:} We aim to derive an upper bound \(U\) such that, for any function \(r(x, y)\) satisfying certain regularity conditions and any finite \(\beta > 0\), the following inequality holds:
\begin{equation}
\mathbb{E}_{y \sim p(\cdot \mid x)}\left[\exp\left(\frac{r(x, y)}{\beta}\right)\right]^\beta - \exp\left(\mathbb{E}_{y \sim p(\cdot \mid x)}[r(x, y)]\right)
\le U\bigl(r(x, y), \beta\bigr).
\end{equation}
Due to the strict convexity of the exponential function (and hence the strictness of Jensen's inequality), the left-hand side is strictly positive unless \(r(x, y)\) is constant almost everywhere under \(p(\cdot \mid x)\) which is already proved in Lemma \ref{lemma:unbiased_estimation}. We now present an upper bound derived using a second-order Taylor expansion. Define:
\begin{equation}
m = \mathbb{E}_{y \sim p(\cdot \mid x)}[r(x, y)], \quad \text{and} \quad V = \operatorname{Var}_{y \sim p(\cdot \mid x)}[r(x, y)].
\end{equation}
For any fixed \(y\), when \(\beta\) is sufficiently large (i.e., when \(\frac{r(x, y)}{\beta}\) is small), we expand the exponential function via a Taylor series:
\begin{equation}
\exp\left(\frac{r(x, y)}{\beta}\right)
= 1 + \frac{r(x, y)}{\beta} + \frac{r(x, y)^2}{2\beta^2} + O\left(\frac{1}{\beta^3}\right).
\end{equation}
Taking expectation with respect to \(y \sim p(\cdot \mid x)\), we obtain:
\begin{equation}
\mathbb{E}\left[\exp\left(\frac{r(x, y)}{\beta}\right)\right]
= 1 + \frac{m}{\beta} + \frac{\mathbb{E}[r(x, y)^2]}{2\beta^2} + O\left(\frac{1}{\beta^3}\right).
\end{equation}
Noting that
\begin{equation}
\mathbb{E}[r(x, y)^2] = V + m^2,
\end{equation}
we can rewrite the expansion as:
\begin{equation}
\mathbb{E}\left[\exp\left(\frac{r(x, y)}{\beta}\right)\right]
= 1 + \frac{m}{\beta} + \frac{m^2 + V}{2\beta^2} + O\left(\frac{1}{\beta^3}\right).
\end{equation}

Next, based on Lemma \ref{lemma:talor_equivalence}, with an error of order $O(\beta^{-2})$:
\begin{equation}
\left(1 + \frac{a}{\beta} + \frac{b}{\beta^2}\right)^\beta = \exp\left(a + \frac{b - \frac{a^2}{2}}{\beta}\right),
\end{equation}
and substitute \(a = m\), \(b = \frac{m^2 + V}{2}\), yielding:
\begin{equation}
\left(\mathbb{E}\left[\exp\left(\frac{r(x, y)}{\beta}\right)\right]\right)^\beta
\approx \exp\left(m + \frac{\frac{m^2 + V}{2} - \frac{m^2}{2}}{\beta}\right)
= \exp\left(m + \frac{V}{2\beta}\right).
\end{equation}
Since \(\exp(m) = \exp\left(\mathbb{E}[r(x, y)]\right)\) corresponds to the lower bound in Jensen's inequality, we arrive at the following second-order approximation:
\begin{equation}
\mathbb{E}_{y \sim p(\cdot \mid x)}\left[\exp\left(\frac{r(x, y)}{\beta}\right)\right]^\beta - \exp\left(\mathbb{E}[r(x, y)]\right)
\approx \exp(m)\left(\exp\left(\frac{V}{2\beta}\right) - 1\right).
\end{equation}

In fact, since the higher-order terms in the Taylor expansion are strictly positive (as a consequence of the strict convexity of the exponential function), the true value of the expression is strictly greater than the second-order approximation. Thus, we may write:
\begin{equation}
\mathbb{E}_{y \sim p(\cdot \mid x)}\left[\exp\left(\frac{r(x, y)}{\beta}\right)\right]^\beta - \exp\left(\mathbb{E}[r(x, y)]\right)
\le \exp\left(\mathbb{E}[r(x, y)]\right)\left(\exp\left(\frac{\operatorname{Var}[r(x, y)]}{2\beta}\right) - 1\right).
\end{equation}

Rewrite the above equation:
\begin{equation}
\frac{\mathbb{E}_{y \sim p(\cdot \mid x)}\left[\exp\left(\frac{r(x, y)}{\beta}\right)\right]^\beta}{\exp\left(\mathbb{E}[r(x, y)]\right)} - 1
\le \left(\exp\left(\frac{\operatorname{Var}[r(x, y)]}{2\beta}\right) - 1\right).
\end{equation}

Add 1 to both sides of the above equation and then remove the logarithm, and we have:
\begin{equation}
0\leq\beta\log Z(x) - \bar{r}(x)
\leq \frac{\operatorname{Var}_{y\sim p(\cdot|x)}[r(x,y)]}{2\beta}.
\end{equation}

This yields an upper bound that depends on the distribution $p(\cdot|x)$, reward function \(r(x, y)\) and its mean and variance under $p(\cdot|x)$, and on the parameter \(\beta\). Proof finished. $\hfill\qedsymbol$

\begin{lemma}\label{lemma:talor_equivalence}
With an error of order $O(\beta^{-2})$,
\begin{equation}
    \bigl(1 + \tfrac{a}{\beta} + \tfrac{b}{\beta^2}\bigr)^\beta = \exp\!\Bigl(a \;+\;\frac{b - \tfrac{a^2}{2}}{\beta}\Bigr).
\end{equation}
\end{lemma}

\textbf{Proof:} Define
\begin{equation}
    L \;=\;\ln\Bigl(1 + \tfrac{a}{\beta} + \tfrac{b}{\beta^2}\Bigr)^\beta
\;=\;\beta\,\ln\Bigl(1 + \tfrac{a}{\beta} + \tfrac{b}{\beta^2}\Bigr).
\end{equation}

Since for large $\beta$ the quantities $a/\beta$ and $b/\beta^2$ are small, one may expand the inner logarithm in powers of $1/\beta$. Set $
u \;=\;\frac{a}{\beta} + \frac{b}{\beta^2},
$

and recall the expansion
$$
\ln(1 + u) \;=\; u \;-\;\frac{u^2}{2}\;+\;\frac{u^3}{3}\;-\;\cdots.
$$

Truncating after the quadratic term (since higher terms will contribute only $O(\beta^{-2})$ or smaller to $L$), we have
$$
\ln\Bigl(1 + \tfrac{a}{\beta} + \tfrac{b}{\beta^2}\Bigr)
\;\approx\;
\frac{a}{\beta} + \frac{b}{\beta^2}
\;-\;\frac{1}{2}\Bigl(\frac{a}{\beta} + \frac{b}{\beta^2}\Bigr)^2.
$$

Then multiply the expansion by $\beta$:
$$
L
\;\approx\;
\beta\Bigl(\frac{a}{\beta} + \frac{b}{\beta^2}\Bigr)
\;-\;\frac{\beta}{2}\Bigl(\frac{a^2}{\beta^2} + \frac{2ab}{\beta^3} + \frac{b^2}{\beta^4}\Bigr)
\;=\;
a \;+\;\frac{b}{\beta}
\;-\;\frac{a^2}{2\beta}
\;+\;O(\beta^{-2}).
$$

Thus,
$$
L \;=\; a \;+\;\frac{b - \tfrac{a^2}{2}}{\beta}
\;+\;O\bigl(\beta^{-2}\bigr).
$$

Recalling that $(1 + a/\beta + b/\beta^2)^\beta = e^L$, we conclude that with an error of order $O(\beta^{-2})$,
\begin{equation}
    \bigl(1 + \tfrac{a}{\beta} + \tfrac{b}{\beta^2}\bigr)^\beta = \exp\!\Bigl(a \;+\;\frac{b - \tfrac{a^2}{2}}{\beta}\Bigr),
\end{equation}
Proof finished. $\hfill\qedsymbol$

\section{Proof of Theorem \ref{theorem:var}}\label{proof:var}
\textbf{Theorem \ref{theorem:var}:} Denote $\bar{r}(x)=\mathbb{E}_{y\sim \pi_{\theta_{\text{old}}}(\cdot|x)}[r(x,y)]$. For $\nabla_\theta \mathcal{L}_{\text{DGRO}}(\theta)$, we have:
    \begin{equation}
    \begin{aligned}
        \nabla_\theta \mathcal{L}_{\text{DGRO}}(\theta) = &-\mathbb{E}_{x \sim \mathcal{D}}\underbrace{\mathbb{E}_{y\sim \pi_{\theta_{\text{old}}}(\cdot|x)}\left[\beta_1\left(r\left(x, y\right)-\bar{r}(x)\right)\nabla_\theta \log \pi_\theta\left(y \mid x\right)\right]}_{\text{PG}(x,\theta)}\\
        &\quad\quad\quad\quad\quad-\mathbb{E}_{x \sim \mathcal{D}}\underbrace{\mathbb{E}_{y\sim \pi_{\theta_{\text{old}}}(\cdot|x)}\left[{\beta_2} \nabla_\theta\left(\log \frac{\pi_\theta\left(y \mid x\right)}{\pi_{\theta_{\text{old}}}\left(y \mid x\right)}\right)^2\right]}_{\text{Normaliser}(x,\theta)}.
    \end{aligned}
    \end{equation}
    Denote $\operatorname{Var}_{y\sim \pi_{\theta_{\text{old}}}(\cdot|x)}[r(x,y)]=\mathbb{E}_{y\sim \pi_{\theta_{\text{old}}}(\cdot|x)}[(r(x,y)-\bar{r}(x))^2]$. Assume that $r(x,y)$ is bounded by $r_{\text{max}}$ and has finite variance, $|\log\frac{\pi_\theta(y|x)}{\pi_{\theta_{\text{old}}}(y|x)}| \leq \eta$ and $\pi_\theta(y|x)$ is parameterized by $\text{Softmax}(f({x},{y};\theta))$, the following property holds:
    \begin{equation}
    \begin{aligned}
        &\left\|\text{PG}(x,\theta)\right\| \leq  (4r_{\text{max}}+2)N\beta_1\gamma(x ; \theta)\operatorname{Var}_{y \sim \pi_{\theta_{\text{old}}}(\cdot \mid x)}[r(x, y)]^{\frac{1}{3}}\\
        &\left\|\text{Normaliser}(x,\theta)\right\|\leq 4N\eta\beta_2\gamma(x ; \theta),
    \end{aligned}
    \end{equation}
    where $\gamma(x;\theta):=\max_{y \in \mathcal{Y},\ n \in {1, \ldots, N}} || J_{f(x, y_{\leq n-1}; \theta)} ||_2$, with $y_{\leq n-1}$ denoting the first $n$ tokens of the response $y$ and $J_{f({x}, {y}_{\leq n-1} ; \theta)}$ is the Jacobian of $f\left({x}, {y}_{\leq n-1}; \theta\right)$ with respect to $\theta$, and $||\cdot||$ and $||\cdot||_2$ denote the Euclidean and operator norms, respectively.

\subsection{Proof of upper bound of $\left\|\text{PG}(x,\theta)\right\|$}\label{proof:PG_term}
Firstly, we prove the upper bound of Policy Gradient term:
\begin{equation}
    \left\|\text{PG}(x,\theta)\right\| \leq (4r_{\text{max}}+2)N\beta_1\gamma(x ; \theta)\operatorname{Var}_{y \sim \pi_{\theta_{\text{old}}}(\cdot \mid x)}[r(x, y)]^{\frac{1}{3}}.
\end{equation}
In the proof of the upper bound on $||\text{PG}(\theta)||$, we follow the general strategy outlined in \cite{razin2023vanishing}, which considers two separate cases and leverages techniques such as Chebyshev’s inequality to relate the bound to the variance of the reward function. However, unlike \cite{razin2023vanishing}, where the target distribution for gradient estimation matches the sampling distribution, our setting is off-policy, meaning the target and sampling distributions differ. Under this off-policy scenario, we show that the norm of the policy gradient is upper bounded by the variance of the reward function under the sampling distribution, rather than the target distribution. Below is proof details.

Because we assumed that $\pi_\theta(y|x)$ is parameterized by $\text{Softmax}(f({x}, {y} ; \theta))$, we have the following derivation for $\nabla_\theta \log \pi_\theta\left(y \mid x\right)$:
\begin{equation}
\begin{aligned}
&\nabla_\theta \log \pi_\theta\left(y \mid x\right) =\sum_{n=1}^{N} \nabla_\theta \log  \pi_\theta\left(y_n \mid x, y_{\leq n-1}\right) \\
=&\sum_{n=1}^{N} \nabla_\theta \log \operatorname{softmax}\left(f(x, y_{\leq n-1} ; \theta)\right)_{y_n} = \sum_{n=1}^{N} J_{f(x, y_{\leq n-1} ; \theta)}^{\top}\left(\mathbf{e}_{y_n}-\pi_\theta\left(\cdot \mid x, y_{\leq n-1}\right)\right),
\end{aligned}
\end{equation}
where $N$ is the response length, $J_{f(x, y_{\leq n-1} ; \theta)}^{\top}$ is the Jacobian of $f\left({x}, {y}_{\leq n-1}; \theta\right)$ and $\mathbf{e}_{y_n}$ is a unit vector where the $y_n$'th entry of $\mathbf{e}_{y_n}$ is 1 and others are 0. 

Let $\mathcal{X}$ and $\mathcal{Y}$ denote the prompt space and the response space, respectively. For $\text{PG}(x,\theta)$, we have:
\begin{equation}
    \begin{aligned}
\text{PG}(x,\theta)=& \mathbb{E}_{y\sim \pi_{\theta_{\text{old}}}(\cdot|x)}\left[\beta_1\left(r\left(x, y\right)-\bar{r}(x)\right)\nabla_\theta \log \pi_\theta\left(y \mid x\right)\right] \\
=&\sum_{y \in \mathcal{Y} } \beta_1\pi_{\theta_{\text{old}}}(y \mid x) r(x, y) \sum_{n=1}^{N} J_{f(x, y_{\leq n-1} ; \theta)}^{\top}\left(\mathbf{e}_{y_n}-\pi_\theta\left(\cdot \mid x, y_{\leq n-1}\right)\right).
    \end{aligned}
\end{equation}

Given $c>0$ which will be determined later, denote by $\mathcal{Y}_c$ the set of outputs whose rewards deviate by more than $c$ from the expected reward, i.e.:
\begin{equation}
    \mathcal{Y}_c:=\left\{y \in \mathcal{Y}:|r(x, y)-\bar{r}(x)|>c\right\}.
\end{equation}

We can write $\text{PG}(x,\theta)$ as follows:
\begin{equation}
    \begin{aligned}
&\text{PG}(x,\theta)\\
=&\underbrace{\sum_{y \in \mathcal{Y}/\mathcal{Y}_c }\beta_1 \pi_{\theta_{\text{old}}}(y \mid x) (r(x, y)-\bar{r}(x)) \sum_{n=1}^{N} J_{f(x, y_{\leq n-1} ; \theta)}^{\top}\left(\mathbf{e}_{y_n}-\pi_\theta\left(\cdot \mid x, y_{\leq n-1}\right)\right) }_{\text{PG}_1(x,\theta)}\\
&+\underbrace{\sum_{y \in \mathcal{Y}_c } \beta_1\pi_{\theta_{\text{old}}}(y \mid x) (r(x, y)-\bar{r}(x)) \sum_{n=1}^{N} J_{f(x, y_{\leq n-1} ; \theta)}^{\top}\left(\mathbf{e}_{y_n}-\pi_\theta\left(\cdot \mid x, y_{\leq n-1}\right)\right)}_{\text{PG}_2(x,\theta)}.
    \end{aligned}
\end{equation}

Notice that $\left\|\mathbf{e}_{y_n}-\pi_\theta\left(\cdot \mid x, y_{\leq n-1}\right)\right\| \leq\left\|\mathbf{e}_{y_n}-\pi_\theta\left(\cdot \mid x, y_{\leq n-1}\right)\right\|_1 \leq 2$, where $\|\cdot\|_1$ denotes the $\ell_1$ norm. Thus, for any $y \in \mathcal{Y}$ and $n \in\left\{1, \ldots, N\right\}$ :
\begin{equation}
    \begin{aligned}
\left\|J_{f(x, y_{\leq n-1} ; \theta)}^{\top}\left(\mathbf{e}_{y_n}-\pi_\theta\left(\cdot \mid x, y_{\leq n-1}\right)\right)\right\| & \leq\left\|J_{f(x, y_{\leq n-1} ; \theta)}^{\top}\right\|_2 \cdot\left\|\left(\mathbf{e}_{y_n}-\pi_\theta\left(\cdot \mid x, y_{\leq n-1}\right)\right)\right\| \\
& \leq 2 \gamma(x ; \theta).
\end{aligned}
\end{equation}
where $\gamma(x;\theta):=\max_{y \in \mathcal{Y},\ n \in {1, \ldots, N}} || J_{f(x, y_{\leq n-1}; \theta)} ||_2$.

For $\text{PG}_1(\theta)$, we have:
\begin{equation}
    ||\text{PG}_1(\theta)|| \leq 2Nc\beta_1 \gamma(x ; \theta)
\end{equation}

Assume that $r(x,y)$ is bound between $[-r_{\text{max}},r_{\text{max}}]$, then $|r(x, y)-\bar{r}(x)|\leq2r_{\text{max}}$. With Chebyshev's inequality we know that:
\begin{equation}
    \pi_{\theta_{\text{old}}}\left(\mathcal{Y}_c \mid x\right) \leq \frac{\operatorname{Var}_{y \sim \pi_{\theta_{\text{old}}}(\cdot \mid x)}[r(x, y)]}{c^2}.
\end{equation}

For $\text{PG}_2(\theta)$, we have:
\begin{equation}
    ||\text{PG}_2(\theta)|| \leq 4N\gamma(x ; \theta)r_{\text{max}}\frac{\operatorname{Var}_{y \sim \pi_{\theta_{\text{old}}}(\cdot \mid x)}[r(x, y)]}{c^2}.
\end{equation}

Let $c=\operatorname{Var}_{y \sim \pi_{\theta_{\text{old}}}(\cdot \mid x)}[r(x, y)]^{\frac{1}{3}}$, then 
\begin{equation}
    ||\text{PG}(x,\theta)||\leq (4r_{\text{max}}+2)N\beta_1\gamma(x ; \theta)\operatorname{Var}_{y \sim \pi_{\theta_{\text{old}}}(\cdot \mid x)}[r(x, y)]^{\frac{1}{3}}.
\end{equation}
Proof finished. $\hfill\qedsymbol$ 

\subsection{Proof of upper bound of $\|\text{Normaliser}(x,\theta)\|$}
\begin{equation}
    \begin{aligned}
&\text{Normaliser}(x,\theta)\\
=& \mathbb{E}_{y\sim \pi_{\theta_{\text{old}}}(\cdot|x)}\left[{\beta_2} \nabla_\theta\left(\log \frac{\pi_\theta\left(y \mid x\right)}{\pi_{\theta_{\text{old}}}\left(y \mid x\right)}\right)^2\right] \\
=&\sum_{y \in \mathcal{Y} } 2\beta_2\pi_{\theta_{\text{old}}}\left(y \mid x\right)\log \frac{\pi_\theta\left(y \mid x\right)}{\pi_{\theta_{\text{old}}}\left(y \mid x\right)}\nabla_\theta \log \pi_\theta\left(y \mid x\right)  \\
=&\sum_{y \in \mathcal{Y} } 2\beta_2\pi_{\theta_{\text{old}}}\left(y \mid x\right)\log \frac{\pi_\theta\left(y \mid x\right)}{\pi_{\theta_{\text{old}}}\left(y \mid x\right)} \sum_{n=1}^{N} J_{f(x, y_{\leq n-1} ; \theta)}^{\top}\left(\mathbf{e}_{y_n}-\pi_\theta\left(\cdot \mid x, y_{\leq n-1}\right)\right).
    \end{aligned}
\end{equation}

Note that we assume $|\log\frac{\pi_\theta(y|x)}{\pi_{\theta_{\text{old}}}(y|x)}| \leq \eta$, according to the conclusions in Appendix \ref{proof:PG_term}, we have:
\begin{equation}
    ||\text{Normaliser}(x,\theta)||\leq 4N\eta\beta_2\gamma(x ; \theta).
\end{equation}
Proof finished. $\hfill\qedsymbol$ 






\section{Experimental Details}
\label{sec:exp_details}

\subsection{Reward Design}
\label{sec:reward_design}
                                 
We consider three reward types, that is, the format score $r_f$, correct reward $r_{corr}$ and complete reward $r_{com}$, each of which equals to $1$ if the corresponding type is achieved, otherwise is $-1$. We control the reward variance by adjusting the range.
$$r = c_f \cdot r_f + c_{\text{corr}} \cdot r_{\text{corr}} + c_{\text{com}} \cdot r_{\text{com}}\,.$$
For the three reward settings with different variances in the ablation experiment, the parameters of format reward $c_f$, correct reward $c_{corr}$ and complete reward $c_{com}$ are defined as follows, and the final corresponding scores are in Table \ref{table:reward_design}.

(1) $r\in [-1, 1]$: \( c_f = 0.4 \), \( c_{\text{corr}} = 0.5 \), and \( c_{\text{com}} = 0.1 \);

(2) $r\in [-3, 3]$: \( c_f = 1.0 \), \( c_{\text{corr}} = 1.75 \), and \( c_{\text{com}} = 0.25 \);

(3) $r\in [-5, 5]$: \( c_f = 2.0 \), \( c_{\text{corr}} = 2.5 \), and \( c_{\text{com}} = 0.5 \);

\begin{table}[h]
\centering
\caption{Reward Values under Different Reward Scaling Schemes.}
\begin{tabular}{lcccc}
\toprule
\textbf{Reward Range} &
\makecell[c]{Correct format,\\ correct answer, complete} &
\makecell[c]{Correct format,\\ incorrect answer} &
\makecell[c]{Correct format,\\ incomplete answer} &
\makecell[c]{Incorrect\\ format} \\
\midrule
$r \in [-1,\ 1]$ & 1.0 & 0.0 & -0.2 & -1.0 \\
$r \in [-3,\ 3]$ & 3.0 & -0.5 & -1.0 & -3.0 \\
$r \in [-5,\ 5]$ & 5.0 & 0.0 & -1.0 & -5.0 \\
\bottomrule
\end{tabular}
\label{table:reward_design}
\end{table}

\subsection{More Experimental Details}
\label{sec:more_exp}

\paragraph{Hyperparameters.} Table \ref{tab:hyperpameters} shows the parameters in our practical experiments.
\begin{table}[htbp]
\centering
\caption{Hyperparameters for Logic and Math Tasks}
\resizebox{1\linewidth}{!}{
\begin{tabular}{@{}lll@{}}
\toprule
\textbf{Category} & \textbf{Logic} & \textbf{Math} \\
\midrule
\texttt{n\_gpus\_per\_node} & 4 & 8 \\
\texttt{GPU type} & 4$\times$A100 & 8$\times$MI300 \\
\texttt{Base\_Model} & Qwen2.5-7B-Instruct-1M & DeepSeek-R1-Distill-Qwen-7B \\
\texttt{$(\beta_1, \beta_2)$} & $(1, 0.1)$ & $(1, 0.1)$ \\
\texttt{temperature} & 1.0 & 1.0 \\
\texttt{rollout\_n} & 8 & 8 \\
\texttt{lr} & 4e-7 & 4e-7 \\
\texttt{train\_batch\_size} & 4 & 8 \\
\texttt{val\_batch\_size} & 8 & 8 \\
\texttt{max\_prompt\_length} & 400 & $2,048$ \\
\texttt{ppo\_mini\_batch\_size} & 128 & 256 \\
\texttt{ppo\_micro\_batch\_size} & 32 & 64 \\
\texttt{log\_prob\_micro\_batch\_size} & 160 & 160 \\
\texttt{max\_response\_length} & $4,096$ & $8,192$ \\
\texttt{ppo\_max\_token\_len\_per\_gpu} & $24,576$ & $18,432$ \\
\bottomrule
\end{tabular}}
\label{tab:hyperpameters}
\end{table}

\begin{figure}[htbp]
    \centering
    
    \begin{subfigure}[t]{0.4\textwidth}
        \centering
        \includegraphics[width=\textwidth]{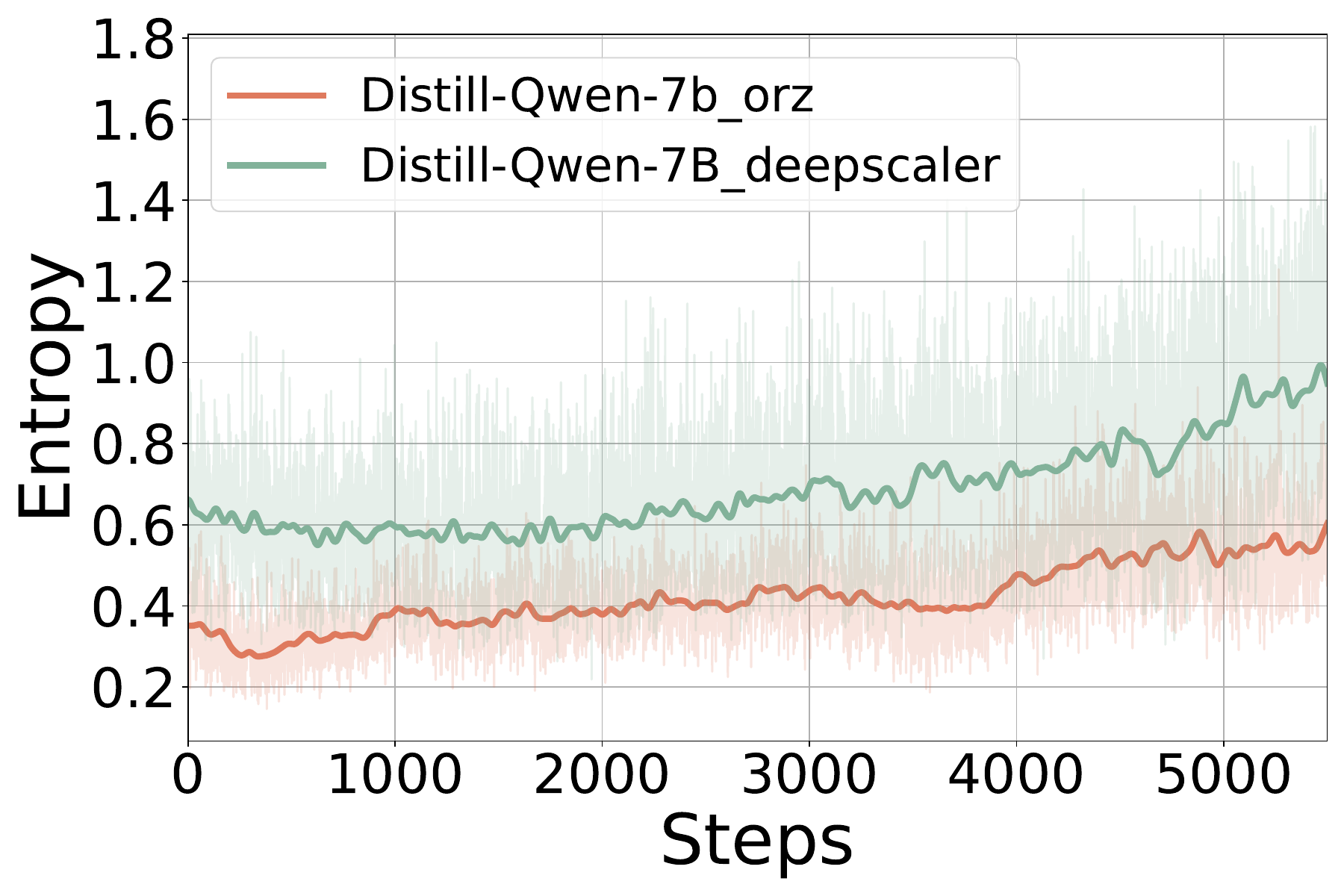}
    \end{subfigure}
    \caption{Entropy during training for math reasoning. The base model is \texttt{DeepSeek-R1} \texttt{-Distill-Qwen-7B}, and the training dataset is orz and deepscaler respectively.}
    \label{fig:math}
\end{figure}

\paragraph{More Results.}
For simple reasoning tasks such as logic task reasoning, entropy typically converges rapidly, which can be seen in ablation study. However, for more complex mathematical reasoning, continued entropy increase is observed in Figure \ref{fig:math}, providing explanation for exploration.

\newpage
\section{Case Study}\label{sec:case_study}
Here we show part of the case study. We use different colored fonts to enhance readability, where \textcolor{sys_prompt}{green fonts represent system prompts}, \textcolor{question}{purple fonts represent questions}, \textcolor{CoT}{orange fonts represent Chain of Thought}, and \textcolor{answer}{blue fonts represent answers}.
\begin{tcolorbox}[
    colframe=c1, 
    colback=blue!5!white, 
    coltitle=white, 
    fonttitle=\bfseries, 
    title=Case Study for model \texttt{Qwen2.5-7B-Instruct-1M} + DGRO on K\& K, 
    arc=3mm, 
    boxrule=0.5mm 
]
\textbf{Question}: A very special island is inhabited only by knights and knaves. Knights always tell the truth, and knaves always lie. You meet 3 inhabitants: Samuel, Ella, and Logan. Samuel commented, "Logan is a knight if and only if Logan is a knave". Ella remarked, "Ella is a knight". In Logan's words: "If Logan is a knight then Ella is a knave". So who is a knight and who is a knave?\\

\textbf{Answer}: (1) Logan is a knight (2) Ella is a knave (3) Samuel is a knave. \\

\textcolor{sys_prompt}{<|im\_start|>system\textbackslash nYou are a helpful assistant.<|im\_end|>\textbackslash n<|im\_start|>user\textbackslash n<|im\_start|> system\textbackslash nYou are a helpful assistant. The assistant first thinks about the reasoning process in the mind and then provides the user with the answer. The reasoning process and answer are enclosed within <think> </think> and<answer> </answer> tags, respectively, i.e., <think> reasoning process here </think><answer> answer here </answer>.  Now the user asks you to solve a logical reasoning problem. After thinking, when you finally reach a conclusion, clearly state the identity of each character within <answer> </answer> tags. i.e., <answer> (1) Zoey is a knight\textbackslash n(2) ... </answer>.\textbackslash n<|im\_end|>\textbackslash n}\textcolor{question}{<|im\_start|>user\textbackslash nA very special island is inhabited only by knights and knaves. Knights always tell the truth, and knaves always lie. You meet 4 inhabitants: Henry, Jack, Amelia, and Evelyn. Henry was heard saying, "Jack is a knave if and only if Henry is a knight". Jack commented, "Henry is not a knight". Amelia expressed that Evelyn is a knave. In a statement by Evelyn: "Jack is a knight and Henry is a knight". So who is a knight and who is a knave?\textbackslash n<|im\_end|>\textbackslash n<|im\_start|>assistant\textbackslash n<think>}\textcolor{CoT}{<|im\_end|>\textbackslash n<|im\_start|>assistant\textbackslash n 
Let\'s analyze the statements step by step.\textbackslash n\textbackslash n 1. Samuel\'s statement: "Logan is a knight if and only if Logan is a knave".\textbackslash n 2. Ella\'s statement: "Ella is a knight".\textbackslash n 3. Logan\'s statement: "If Logan is a knight, then Ella is a knave".\textbackslash n\textbackslash n First, consider the nature of "Logan is a knight" and "Logan is a knave". These two statements are mutually exclusive: Logan can either be a knight or a knave, but not both.\textbackslash n\textbackslash n **Case 1: Assume Logan is a knight.**\textbackslash n\textbackslash n- Samuel\'s statement: "Logan is a knight if and only if Logan is a knave". Since Logan is a knight, "Logan is a knight" is true, and "Logan is a knave" is false. Samuel\'s statement becomes "true if and only if false", which is "true if and only if false". This is a false statement because "true" is not equal to "false". Therefore, Samuel is a knave.\textbackslash n\textbackslash n- Ella\'s statement: "Ella is a knight". To determine the truth value of Ella\'s statement, we need to know the truth value of Ella.\textbackslash n\textbackslash n - Logan\'s statement: "If Logan is a knight, then Ella is a knave". Since Logan is a knight, the antecedent "Logan is a knight" is true. Logan\'s statement depends on the truth value of Ella.\textbackslash n\textbackslash n**Sub-case 1.1: Assume Ella is a knight.**\textbackslash n\textbackslash n- Ella\'s statement: "Ella is a knight". Since Ella is a knight, her statement "Ella is a knight" is true. Therefore, Ella is a knight.\textbackslash n\textbackslash n- Logan\'s statement: "If Logan is a knight, then Ella is a knave". Since Logan is a knight, the antecedent "Logan is a knight" is true. However, Ella is a knight, so the consequent "Ella is a knave" is false. Therefore, Logan\'s statement is false, meaning Logan is a knave. This creates a contradiction because we previously assumed Logan is a knight. Therefore, Ella cannot be a knight.\textbackslash n\textbackslash n**Sub-case 1.2: Assume Ella is a knave.**\textbackslash n\textbackslash n- Ella\'s statement: "Ella is a knight". Since Ella is a knave, her statement "Ella is a knight" is false. Therefore, Ella\'s statement is false, meaning Ella is a knave.\textbackslash n\textbackslash n - Logan\'s statement: "If Logan is a knight, then Ella is a knave". Since Logan is a knight, the antecedent "Logan is a knight" is true. Ella is a knave, so the consequent "Ella is a knave" is true. Therefore, Logan\'s statement is true, meaning Logan is a knight.\textbackslash n\textbackslash n- Samuel\'s statement: "Logan is a knight if and only if Logan is a knave". Since Logan is a knight, "Logan is a knight" is true, and "Logan is a knave" is false. Samuel\'s statement becomes "true if and only if false", which is "true if and only if false". This is a false statement because "true" is not equal to "false". Therefore, Samuel is a knave.\textbackslash n\textbackslash n Let\'s verify the statements:\textbackslash n\textbackslash n - Ella\'s statement: "Ella is }
\end{tcolorbox}

\begin{tcolorbox}[
    colframe=c1, 
    colback=blue!5!white, 
    coltitle=white, 
    fonttitle=\bfseries, 
    title=Case Study for DGRO on Logic (Continue), 
    arc=3mm, 
    boxrule=0.5mm 
]
\textcolor{CoT}{
a knight". Since Ella is a knave, her statement "Ella is a knight" is false. Therefore, Ella is a knave.\textbackslash n\textbackslash n- Logan\'s statement: "If Logan is a knight, then Ella is a knave". Since Logan is a knight, the antecedent "Logan is a knight" is true. Ella is a knave, so the consequent "Ella is a knave" is true. Therefore, Logan\'s statement is true, meaning Logan is a knight.\textbackslash n\textbackslash n - Samuel\'s statement: "Logan is a knight if and only if Logan is a knave". Since Logan is a knight, "Logan is a knight" is true, and "Logan is a knave" is false. Samuel\'s statement becomes "true if and only if false", which is false. Therefore, Samuel is a knave.\textbackslash n\textbackslash n The solution is consistent. Therefore, the answer is:\textbackslash n\textbackslash n - Logan is a knight.\textbackslash n- Ella is a knave.\textbackslash n- Samuel is a knave. </think> <answer>(1) Logan is a knight\textbackslash n(2) Ella is a knave\textbackslash n (3) Samuel is a knave</answer>.
\textbackslash n</answer><|im\_end|>} \textcolor{answer}{The answer is (1) Logan is a knight (2) Ella is a knave (3) Samuel is a knave.}
\end{tcolorbox}

\begin{tcolorbox}[
    colframe=c1, 
    colback=blue!5!white, 
    coltitle=white, 
    fonttitle=\bfseries, 
    title=Case Study for model \texttt{DeepSeek-R1-Distill-Qwen-7B(8k)+DGRO on Math}, 
    arc=3mm, 
    boxrule=0.5mm 
]
\textbf{Question}:
Jen enters a lottery by picking $4$ distinct numbers from $S=\{1,2,3,\cdots,9,10\}.$ $4$ numbers are randomly chosen from $S.$ She wins a prize if at least two of her numbers were $2$ of the randomly chosen numbers, and wins the grand prize if all four of her numbers were the randomly chosen numbers. The probability of her winning the grand prize given that she won a prize is $\frac{m}{n}$ where $m$ and $n$ are relatively prime positive integers. Find $m+n$.

\textbf{Answer}: $116$

\textcolor{question}{<|begin\_of\_sentence|><|User|> Jen enters a lottery by picking $4$ distinct numbers from $S=\{1,2,3,\ cdots,9,10\}.$ $4$ numbers are randomly chosen from $S.$ She wins a prize if at least two of her numbers were $2$ of the randomly chosen numbers, and wins the grand prize if all four of her numbers were the randomly chosen numbers. The probability of her winning the grand prize given that she won a prize is \textbackslash frac\{m\}\{n\} where $m$ and $n$ are relatively prime positive integers. Find $m+n$.}
\textcolor{sys_prompt}{Let's think step by step and output the final answer within \textbackslash\textbackslash boxed\{\}.}\textcolor{CoT}{<|Assistant|><think>\textbackslash To solve this problem, we need to calculate the probability of Jen winning the grand prize given that she won a prize. Let's break this down step by step.\textbackslash n\textbackslash n First, we need to determine the total number of ways to pick 4 numbers out of 10. This is given by the combination formula \textbackslash( \textbackslash binom{10}{4} \textbackslash):\textbackslash n\textbackslash n\textbackslash[\textbackslash n\textbackslash binom{10}{4} = \textbackslash frac{10!}{4!6!} = 210\textbackslash n\textbackslash]\textbackslash n\textbackslash n Now, let's denote Jen's chosen numbers as \textbackslash ( A = \textbackslash{$a_1, a_2, a_3, a_4$\textbackslash} \textbackslash) and the lottery's chosen numbers as \textbackslash( B = \textbackslash{$b_1, b_2, b_3, b_4$\textbackslash} \textbackslash).\textbackslash n\textbackslash n We need to find the probability that Jen wins the grand prize given that she won a prize. This means we need to calculate the conditional probability \textbackslash( P(\textbackslash text{Grand Prize} \textbackslash mid \textbackslash text{Prize}) \textbackslash).\textbackslash n\textbackslash n The probability of Jen winning the grand prize (i.e., \textbackslash( A = B \textbackslash)) is simply:\textbackslash n\textbackslash n \textbackslash[\textbackslash n P(\textbackslash text{Grand Prize}) = \textbackslash frac{1}{\textbackslash binom{10}{4}} = \textbackslash frac{1}{210}\textbackslash n\textbackslash]\textbackslash n\textbackslash n Now, we need to find the probability of Jen winning a prize (i.e., at least 2 of her numbers match the lottery's numbers). To do this, we need to calculate the probability of the complementary event (i.e., winning no prize) and subtract it from 1.\textbackslash n\textbackslash n The number of ways to choose 4 numbers from 10 such that none or only one of them matches Jen's numbers can be calculated as follows:\textbackslash n\textbackslash n 1. **No match**: All 4 numbers are chosen from the remaining 6 numbers (those not in Jen's set). There are \textbackslash( \textbackslash binom{6}{4} \textbackslash) ways to do this:\textbackslash n\textbackslash n \textbackslash[\textbackslash n\textbackslash binom{6}{4} = \textbackslash binom{6}{2} = 15\textbackslash n\textbackslash]\textbackslash n\textbackslash n 2. **Exactly one match**: Choose 1 number from Jen's 4 numbers and 3 numbers from the remaining 6 numbers. There are \textbackslash( \textbackslash binom{4}{1} \textbackslash binom{6}{3} \textbackslash) ways to do this:\textbackslash n\textbackslash n \textbackslash[\textbackslash n\textbackslash binom{4}{1} \textbackslash binom{6}{3} = 4 \textbackslash times 20 = 80\textbackslash n\textbackslash]\textbackslash n\textbackslash n So, the total number of ways to have no match or exactly one match is:\textbackslash n\textbackslash n \textbackslash[\textbackslash n 15 + 80 = 95\textbackslash n\textbackslash]\textbackslash n\textbackslash n Thus, the number of favorable outcomes for winning a prize is:\textbackslash n\textbackslash n \textbackslash[\textbackslash n210 - 95 = 115\textbackslash n\textbackslash]\textbackslash n\textbackslash n So, the probability of Jen winning a prize is:\textbackslash n\textbackslash n \textbackslash[\textbackslash n P(\textbackslash text{Prize}) = \textbackslash frac{115}{210} = \textbackslash frac{23}{42}\textbackslash n\textbackslash]\textbackslash n\textbackslash n Now, we need to find the conditional probability \textbackslash( P(\textbackslash text{Grand Prize} \textbackslash mid \textbackslash text{Prize}) \textbackslash):\textbackslash n\textbackslash n \textbackslash[\textbackslash n  P(\textbackslash text{Grand Prize} \textbackslash mid \textbackslash text{Prize}) = \textbackslash frac{P(\textbackslash text{Grand Prize})}{P(\textbackslash text{Prize})} = \textbackslash frac{\textbackslash frac{1}{210}}{\textbackslash frac{23}{42}} = \textbackslash frac{1}{210} \textbackslash times \textbackslash frac{42}{23} = \textbackslash frac{42}{210 \textbackslash times 23} = \textbackslash frac{42}{4830} = \textbackslash frac{1}{115}\textbackslash n\textbackslash]\textbackslash n\textbackslash n Thus, the probability of her winning the grand prize given that she won a prize is \textbackslash( \textbackslash frac{1}{115} \textbackslash). Here, \textbackslash( m = 1 \textbackslash) and \textbackslash( n = 115 \textbackslash), and since 1 and 115 are relatively prime, we have \textbackslash( m + n = 1 + 115 = 116 \textbackslash).\textbackslash n\textbackslash n Therefore, the final answer is:\textbackslash n\textbackslash n \textbackslash[\textbackslash n\textbackslash boxed{116}\textbackslash n\textbackslash].}
\textcolor{answer}{The final answer is: \textbackslash\textbackslash boxed\{116\}}
\end{tcolorbox}


\end{document}